%% file: iclr2027_conference.tex
\documentclass{article} 
\usepackage{iclr2027_conference,times}

\input{math_commands.tex}

\usepackage{hyperref}
\usepackage{url}

\usepackage{graphicx}
\usepackage[T1]{fontenc}
\usepackage{booktabs,multirow,calc}
\usepackage{colortbl}
\usepackage{caption}
\usepackage{array}
\usepackage{makecell}

\newlength{\iclrfigonecell}
\newlength{\iclrfigonerowwidth}
\newlength{\iclrfigonegap}
\newlength{\iclrfigoneheaderheight}

\title{Mitigating Sequential Reappearance in \\ Diffusion Data-Point Unlearning}

\author{\textbf{Donghyun Kim\thanks{The first four authors contributed equally.},
Taehyuk Lee\footnotemark[1],
Jinyeong Kim\footnotemark[1],
Youngmin Oh\footnotemark[1],} \\
\textbf{Dohyeong Kim,
Jaehyuk Ryu,
Sangwoo Hong\thanks{Corresponding author.}} \\
Trustworthy Machine Learning Lab, Department of Computer Science, Konkuk University \\
Seoul, Republic of Korea \\
\texttt{\{alphax,matthewsep,jinzero29,peopleoh1204\}@konkuk.ac.kr} \\
\texttt{\{kdh9981,rokmc704,swhong06\}@konkuk.ac.kr}
}
\iclrfinalcopy 
\begin{document}

\maketitle
\lhead{}

\begin{abstract}
Diffusion data-point unlearning is typically evaluated immediately after each deletion, even though subsequent requests may repeatedly update the same model. We identify sequential reappearance, a failure mode in which an instance that is initially judged to be forgotten later returns to the memorized regime without reuse of the deleted data or adversarial fine-tuning. To capture this behavior, we introduce a target-level evaluation protocol that tracks whether each target is forgotten immediately, remains forgotten at the end of the sequence, or reappears during subsequent deletions. We further find that targets that later reappear exhibit greater local recovery accessibility after deletion than targets that remain forgotten. 
We characterize this vulnerability as recovery accessibility and show that this post-deletion local geometry is associated with subsequent reappearance risk.
Motivated by this finding, we propose a recovery-guided unlearning method that iteratively identifies the most recoverable region around each target and extends the deletion objective to that region. Experiments show that our method improves forgetting persistence across long deletion sequences while maintaining competitive generation quality, highlighting the importance of evaluating diffusion unlearning beyond immediate deletion efficacy.
\end{abstract}

\section{Introduction}
\begin{figure*}[t]
\centering

\begin{minipage}[t]{0.67\textwidth}
\centering
\vspace{0pt}

\settowidth{\iclrfigonerowwidth}{%
  \normalfont\footnotesize\bfseries Immediate%
}
\addtolength{\iclrfigonerowwidth}{2pt}
\setlength{\iclrfigonegap}{2pt}

\setlength{\iclrfigonecell}{%
  \dimexpr
  (\linewidth-\iclrfigonerowwidth-5\iclrfigonegap-12pt)/7
  \relax
}
\setlength{\iclrfigoneheaderheight}{2.5\baselineskip}

\newcommand{\iclrfigoneimage}[1]{%
  \parbox[c][\iclrfigonecell][c]{\iclrfigonecell}{%
    \centering
    \includegraphics[
      width=\iclrfigonecell,
      height=\iclrfigonecell,
      keepaspectratio
    ]{#1}%
  }%
}

\newcommand{\iclrfigonepair}[2]{%
  \begin{tabular}[c]{@{}c@{}}
    \iclrfigoneimage{#1}\\
    \noalign{\vskip 4pt}
    \iclrfigoneimage{#2}
  \end{tabular}%
}

\newcommand{\iclrfigonerowlabel}[1]{%
  \parbox[c][\iclrfigonecell][c]{\iclrfigonerowwidth}{%
    \centering
    \normalfont\footnotesize\bfseries
    \shortstack{#1\\Model}%
  }%
}

\newcommand{\iclrfigonerowlabels}{%
  \begin{tabular}[c]{@{}c@{}}
    \iclrfigonerowlabel{Immediate}\\
    \noalign{\vskip 4pt}
    \iclrfigonerowlabel{Final}
  \end{tabular}%
}

\newcommand{\iclrfigoneoriginal}[1]{%
  \parbox[c][\dimexpr 2\iclrfigonecell+4pt\relax][c]
    {\iclrfigonecell}{%
    \centering
    \iclrfigoneimage{#1}%
  }%
}

\newcommand{\iclrfigoneheader}[1]{%
  \parbox[c][\iclrfigoneheaderheight][c]{\iclrfigonecell}{%
    \normalfont\small\bfseries
    \makebox[\iclrfigonecell][c]{\shortstack{#1}}%
  }%
}

\begin{tabular}{
  @{}c
  @{\hspace{\iclrfigonegap}}c
  @{\hspace{6pt}}c
  @{\hspace{6pt}}c
  @{\hspace{\iclrfigonegap}}c
  @{\hspace{\iclrfigonegap}}c
  @{\hspace{\iclrfigonegap}}c
  @{\hspace{\iclrfigonegap}}c@{}
}
  &
  \raisebox{-7pt}[0pt][0pt]{\iclrfigoneheader{Original}}
  &
  \iclrfigoneheader{\textsc{LASTING}\\(Ours)}
  &
  \iclrfigoneheader{SISS}
  &
  \iclrfigoneheader{Erase-\\Diff}
  &
  \iclrfigoneheader{ReTrack}
  &
  \iclrfigoneheader{FU}
  &
  \iclrfigoneheader{Prompt-\\Free}
  \\[3pt]

  \iclrfigonerowlabels
  &
  \iclrfigoneoriginal
    {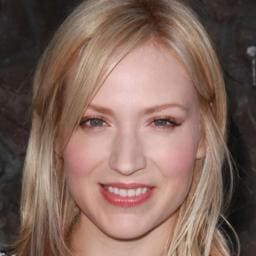}
  &
  \iclrfigonepair
    {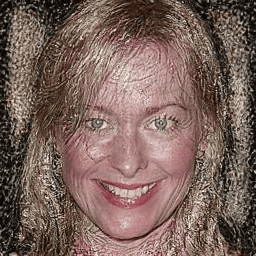}
    {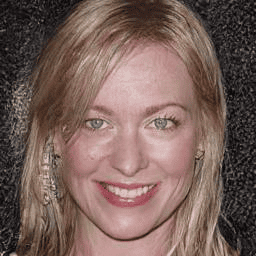}
  &
  \iclrfigonepair
    {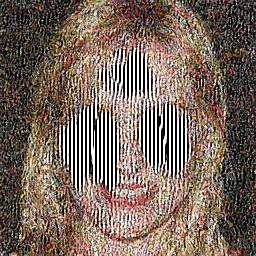}
    {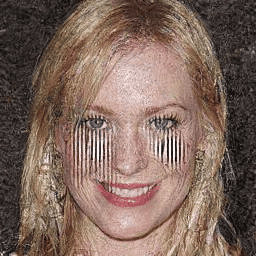}
  &
  \iclrfigonepair
    {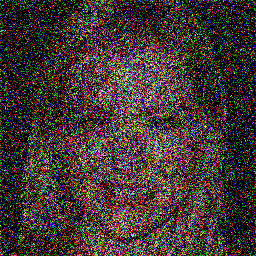}
    {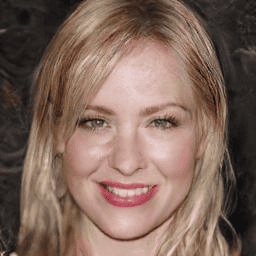}
  &
  \iclrfigonepair
    {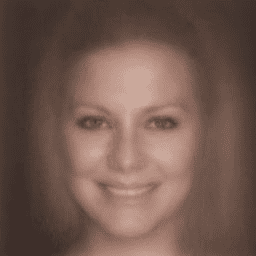}
    {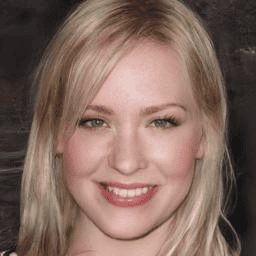}
  &
  \iclrfigonepair
    {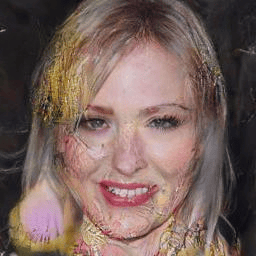}
    {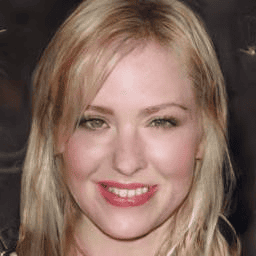}
  &
  \iclrfigonepair
    {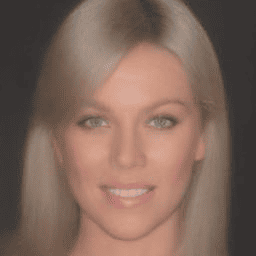}
    {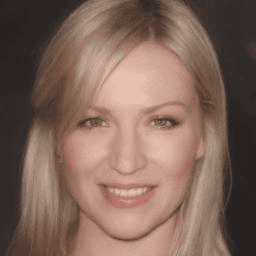}
\end{tabular}

\par\vspace{5pt}
{\normalfont\normalsize (a) Qualitative comparison.\par}
\end{minipage}
\hfill
\begin{minipage}[t]{0.32\textwidth}
\centering
\vspace{0pt}

\includegraphics[
  width=\linewidth,
  keepaspectratio
]{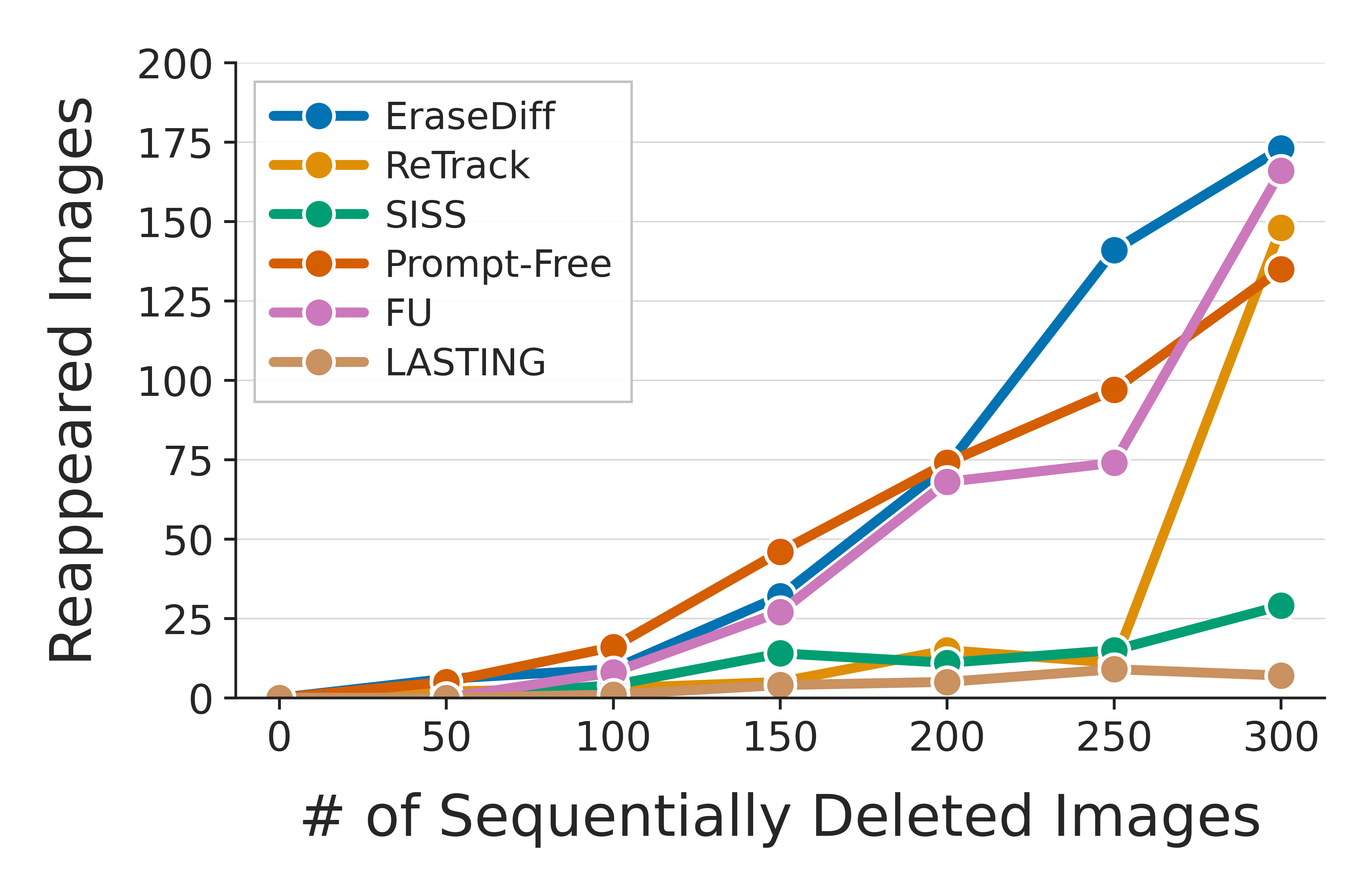}

\par\vspace{5pt}
{\normalfont\normalsize (b) Reappearance counts.\par}
\end{minipage}

\caption{{Sequential reappearance in CelebA-HQ.}
(a) Outputs generated immediately after unlearning each target and
after processing all subsequent deletion requests. Existing methods
can reproduce initially forgotten targets in the final model, whereas
\textsc{LASTING} maintains forgetting after the full sequence of unlearning updates.
(b) Number of reappeared images at each deletion checkpoint.}
\label{fig:sequential-reappearance-overview}

\makeatletter
\begingroup
\edef\@currentlabel{\thefigure a}
\label{fig:sequential-reappearance}
\endgroup
\begingroup
\edef\@currentlabel{\thefigure b}
\label{fig:reappearance-count}
\endgroup
\makeatother

\end{figure*}

Diffusion models can memorize and reproduce individual training examples,
raising privacy and copyright concerns when sensitive data must be
removed~\citep{Somepalli_2023_CVPR,carlini2023extractingtrainingdatadiffusion}.
Diffusion data-point unlearning removes the influence of a specific training
instance without retraining the model from
scratch~\citep{Alberti_2025_ICLR,Shi_2026_AISTATS}. A reliable deletion should
remain effective throughout subsequent updates to the same model. As shown in
Figure~\ref{fig:sequential-reappearance}, however, a target that is forgotten
at its immediate checkpoint can become memorized again after later deletion
requests.

Existing evaluations predominantly assess each target immediately after its
deletion~\citep{Alberti_2025_ICLR,Wu_2025_CVPR,Shi_2026_AISTATS,Park_2026_ICML_FoGen}.
Such stage-local evaluation overlooks the target's behavior under future model
updates and can overestimate deletion reliability. We identify \emph{sequential
reappearance}, in which an initially forgotten target returns to the memorized
regime during ordinary sequential deletion without reuse of the target,
adversarial fine-tuning, or an explicit recovery attack.

To understand this failure, we examine the local denoising-loss geometry around
each target immediately after deletion. We find that targets that later
reappear have greater local recovery accessibility than those that remain forgotten,
even when both satisfy the immediate forgetting criterion. We refer to this
vulnerability as \emph{recovery accessibility}. This finding shows that
reliable unlearning should suppress not only the observed target but also nearby
representations that preserve a route back to memorization.

Motivated by this observation, we propose \emph{Low-loss Adaptive Search and
Targeted Intervention in Neighborhood Geometry} (\textsc{LASTING}).
\textsc{LASTING} repeatedly identifies the most recoverable nearby
representation and uses it to guide the deletion update as the model changes.
Across long sequential deletion trajectories, \textsc{LASTING} substantially improves forgetting persistence while maintaining competitive generation quality. These
results show that successful diffusion unlearning requires forgetting to remain
stable under subsequent deletion requests.

\paragraph{Contributions.}
Our main contributions are summarized as follows:
\begin{itemize}
    \item We introduce a target-level protocol that evaluates both immediate
    and persistent forgetting under sequential diffusion unlearning.
    \item We identify \emph{sequential reappearance} and show that it is
    associated with greater post-deletion recovery accessibility.
    \item We propose \textsc{LASTING}, which adaptively suppresses the most
    locally recoverable representation and substantially improves forgetting
    persistence while maintaining competitive generation quality.
\end{itemize}

\section{Related Work}
\label{sec:related-work}

\subsection{Memorization in Diffusion Models}
Diffusion models can reproduce individual training examples, and extraction
attacks have recovered memorized images at
scale~\citep{Somepalli_2023_CVPR,carlini2023extractingtrainingdatadiffusion}.
Prior work studies how duplication, data coverage, and text conditioning affect
replication and develops detection or inference-time mitigation
methods~\citep{Somepalli_2023_NeurIPS,Wen_2024_ICLR,Merger_2026_ICML_MemFM,Han_2025_NeurIPS}.
Unlike these interventions, diffusion unlearning modifies the model to remove
the influence of selected training data.

\subsection{Diffusion Model Unlearning}
Concept-level unlearning suppresses broad semantic content by editing model
weights, aligning target representations with anchors, or selecting parameters
associated with the target
concept~\citep{Gandikota_2023_ICCV,Kumari_2023_ICCV,Zhang_2024_CVPR,Fan_2024_ICLR,Heng_2023_NeurIPS,Lu_2024_CVPR,Li_2025_ICCV}.
Because deletion is defined semantically, these methods do not precisely remove
an individual training example while preserving related content.

Data-point unlearning instead targets individual examples. EraseDiff redirects
the reverse process, SISS balances forgetting and utility through an
importance-sampled score objective, and ReTrack redirects a target trajectory
toward retain-set neighbors~\citep{Wu_2025_CVPR,Alberti_2025_ICLR,Shi_2026_AISTATS}.
Recent methods formulate unlearning through KL-constrained optimization or use
edited surrogate targets without textual
prompts~\citep{Khalafi_2026_ICML,Lee_2026_CVPR_Workshop_MUV}. These approaches
focus primarily on forgetting immediately after deletion. Our work examines
whether successfully deleted instances remain forgotten as the same model
undergoes subsequent deletion updates.

\subsection{Recovery and Sequential Unlearning}
Recovery after unlearning has mainly been studied under explicit post-unlearning interventions, including fine-tuning on removed data or with unrelated prompts, malicious relearning, and adversarial prompt attacks~\citep{George_2025_CVPR,Gao_2025_ICCV,Zhang_2024_NeurIPS, suriyakumar2024unstable}.
Our setting concerns spontaneous reappearance during ordinary sequential
deletion without additional fine-tuning, recovery attacks, or reuse of the
removed instance.

Sequential diffusion unlearning has largely focused on semantic concepts and
utility preservation across successive removals
~\citep{Lee_2026_ICLR,Thakral_2025_arXiv,George_2026_ECCV}. Closest to our
setting, prior work reports unlearning rebound in sequential style erasure and
forgetting reversal in continual classifier
unlearning~\citep{Zhang_2024_NeurIPS_Unlearncanvas,Park_2026_CVPR_Findings}.
Sequential data-point unlearning in diffusion models remains largely
unexplored, despite its relevance to accumulating copyright and privacy
requests.

\subsection{Local Geometry in Machine Unlearning}
\paragraph{Parameter-space sharpness.}
Loss-landscape sharpness is closely related to generalization and
robustness~\citep{Keskar_2017_ICLR,Dinh_2017_ICML}. Sharpness-Aware
Minimization and Adversarial Weight Perturbation measure loss variation under
parameter perturbations, and recent unlearning methods use similar ideas to
improve stability under relearning
~\citep{Foret_2021_ICLR,Wu_2020_NeurIPS,Tang_2026_ICLR,Malekmohammadi_2025_arXiv,Fan_2025_ICML}.

\paragraph{Data-space local geometry.}
Prior classifier studies examine residual knowledge around perturbed forget
samples or reshape local decision boundaries with adversarial
examples~\citep{Hsu_2025_NeurIPS,Ebrahimpour_2025_ICML}. We instead study the
data-space geometry of the diffusion denoising loss and identify nearby image
representations that remain easy to denoise after deletion. We examine the association between this post-deletion recovery geometry and subsequent reappearance.

\section{Analysis of Sequential Reappearance}
\label{sec:phenomenon}
\subsection{Protocol and Metrics}
\paragraph{Sequential unlearning setup.}
Let $M_{\theta_0}$ be a diffusion model and let
$\mathcal{A}=(a_1,\ldots,a_K)$ be an ordered sequence of deletion targets.
Given retain data $\mathcal{R}$, an unlearning algorithm $\mathcal{U}$ updates
the model at stage $i$ as
\begin{equation}
    \theta_i
    =
    \mathcal{U}\left(\theta_{i-1};a_i,\mathcal{R}\right),
    \qquad i=1,\ldots,K.
    \label{eq:sequential_unlearning}
\end{equation}
For target $a_i$, $\theta_i$ is its \emph{immediate model state}, whereas
$\theta_K$ is the \emph{final model state} after all subsequent requests.
This distinction separates immediate forgetting from forgetting that persists
throughout the deletion trajectory.

\paragraph{Persistence metrics.}
\label{subsec:sequential_reappearance}
Let $s_i(\theta_j)$ denote the SSCD similarity~\citep{Pizzi_2022_CVPR}
between target $a_i$ and its recovery output from $M_{\theta_j}$. We consider
the target forgotten when $s_i(\theta_j)<\tau$ and use $\tau=0.7$ following
prior work~\citep{Somepalli_2023_NeurIPS}. We define Final Success and Final
Reappearance as
\begin{equation}
\begin{aligned}
    \mathrm{FinalSucc.}@\tau
    &=
    \frac{1}{K}
    \sum_{i=1}^{K}
    \mathbf{1}\left[s_i(\theta_K)<\tau\right], \\
    \mathrm{FinalReapp.}@\tau
    &=
    \frac{
        \sum_{i=1}^{K}
        \mathbf{1}\left[
            s_i(\theta_i)<\tau
            \land
            s_i(\theta_K)\geq\tau
        \right]
    }{
        \sum_{i=1}^{K}
        \mathbf{1}\left[
            s_i(\theta_i)<\tau
        \right]
    }.
\end{aligned}
\label{eq:persistence_metrics}
\end{equation}
Final Success measures the fraction of targets forgotten under the final model.
Final Reappearance measures the fraction of immediately forgotten targets that return to the memorized regime under the final model.

\subsection{Reappearance under Sequential Deletion}
We apply this protocol to $K=300$ deletion requests using SISS, EraseDiff,
ReTrack, Prompt-Free, and FU
~\citep{Shi_2026_AISTATS,Wu_2025_CVPR,Alberti_2025_ICLR,Khalafi_2026_ICML,Lee_2026_CVPR_Workshop_MUV}.
Figure~\ref{fig:reappearance-count} shows that previously forgotten targets can
return to the memorized regime as later deletions update the model, confirming
that immediate forgetting does not ensure persistent forgetting.

\paragraph{Recovery accessibility.}
\label{subsec:recovery_geometry}
\label{subsec:loss_sharpness}
To explain this behavior, we examine the local denoising-loss geometry around
each target immediately after deletion. SSCD indicates whether the observed
target is recoverable but does not reveal whether nearby representations remain
easy to denoise. For each target $a_i$, we freeze the immediate model
$\theta_i$, construct a fixed bank $\mathcal{B}_i$ of timesteps and noise
samples, and define the local denoising loss as
\begin{equation}
    L_i(\delta)
    =
    \frac{1}{|\mathcal{B}_i|}
    \sum_{\xi\in\mathcal{B}_i}
    \ell_{\mathrm{diff}}
    \left(\theta_i;a_i+\delta,\xi\right).
    \label{eq:local_recovery_loss}
\end{equation}
Within the bounded data-space neighborhood
$\mathcal{D}_i=\{\delta:\|\delta\|_{\infty}\leq\epsilon, a_i + \delta \in [-1, 1]^d\}$, we approximate a
low-loss neighbor and define \emph{recovery accessibility} as
\begin{equation}
\begin{aligned}
    \widehat{\delta}_i
    &\approx
    \mathop{\arg\min}_{\delta\in\mathcal{D}_i}
    L_i(\delta), \ \ \
    \mathrm{RecAccess}_i
    =
\max\left\{
0,\,
\frac{
L_i(0)-L_i(\hat{\delta}_i)
}{
\max\{L_i(0),\eta\}
}
\right\},
\end{aligned}
\label{eq:local_sharpness}
\end{equation}
where $\widehat{\delta}_i$ is obtained through constrained projected-gradient
search.
Here, ($\eta > 0$) is a small constant for numerical stability.
A larger score indicates that a greater relative reduction in denoising loss is accessible within the prescribed neighborhood.

\begin{figure*}[t]
\centering

\begin{minipage}[t]{0.56\textwidth}
\vspace{0pt}
\centering

\includegraphics[
  width=\linewidth,
  height=1.85in,
  keepaspectratio
]{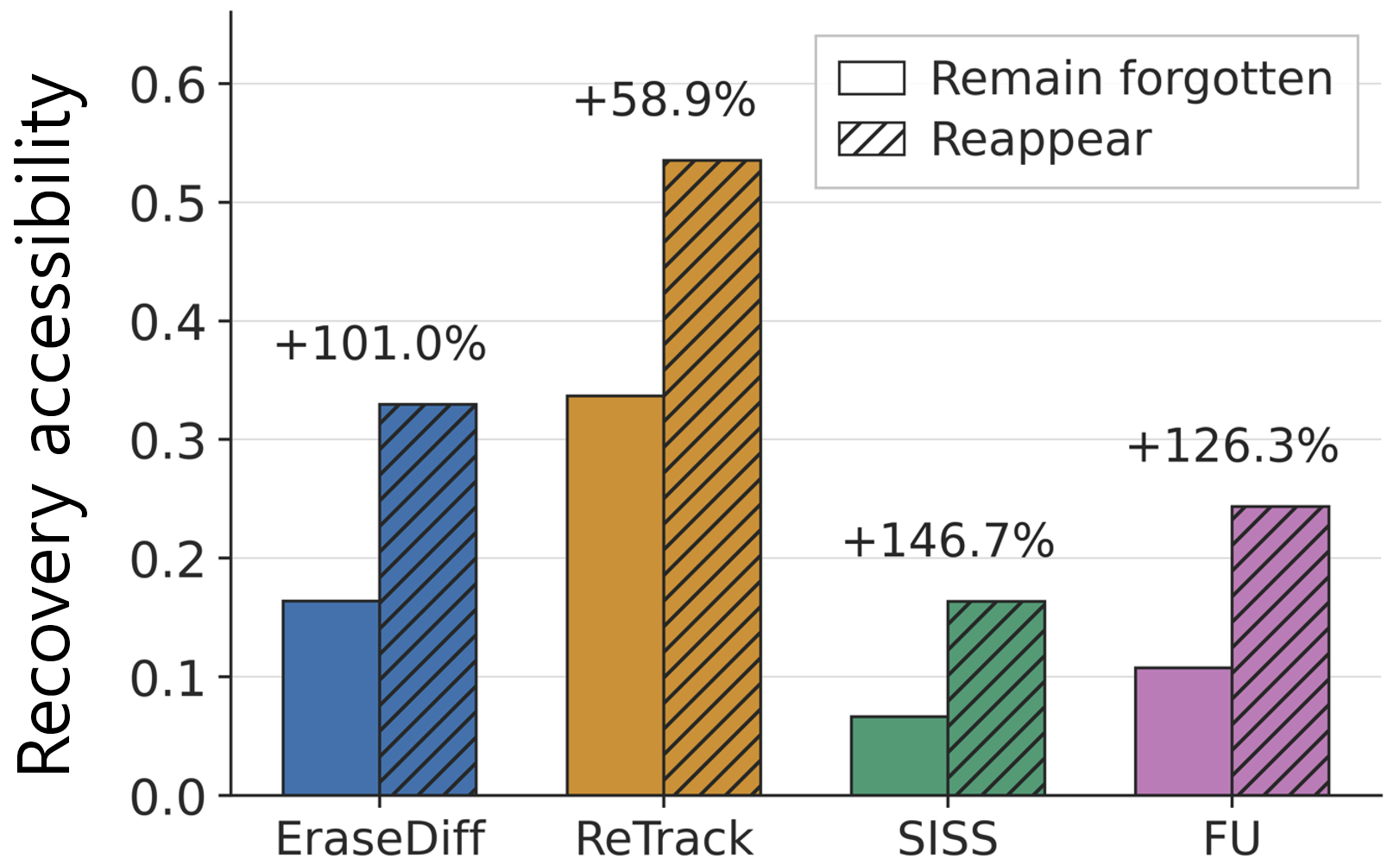}

\caption{{Recovery accessibility by outcome.}
Targets that subsequently reappear exhibit higher recovery accessibility
after the initial unlearning step.}
\label{fig:recovery-sharpness-bar}
\end{minipage}
\hfill
\begin{minipage}[t]{0.40\textwidth}
\vspace{0pt}
\centering

\makeatletter
\def\@captype{table}
\makeatother

\caption{{Association between recovery accessibility and subsequent reappearance.}
$\rho$ denotes Spearman correlation with maximum future SSCD,
partial $\rho$ controls for Immediate SSCD, and AUC denotes AUROC.}
\label{tab:recovery-sharpness-prediction}

\vspace{2pt}
\setlength{\tabcolsep}{4pt}
\renewcommand{\arraystretch}{1.08}

\begin{tabular}{@{}lccc@{}}
\toprule
Method
& $\rho$
& \shortstack{Partial\\$\rho$}
& AUC \\
\midrule
EraseDiff & 0.528 & 0.529 & 0.740 \\
ReTrack   & 0.436 & 0.381 & 0.750 \\
SISS      & 0.452 & 0.234 & 0.707 \\
FU        & 0.358 & 0.349 & 0.669 \\
\bottomrule
\end{tabular}

\end{minipage}

\end{figure*}

\paragraph{Recovery accessibility and reappearance.}
We next examine whether recovery accessibility measured immediately after deletion is associated with subsequent reappearance. To exclude immediate deletion failures, we restrict the analysis to targets satisfying $s_i(\theta_i)<\tau$ and track their outcomes over the following 30 deletion stages. As shown in Figure~\ref{fig:recovery-sharpness-bar}, targets that later reappear consistently exhibit higher recovery accessibility than those that remain forgotten across different unlearning methods.

Table~\ref{tab:recovery-sharpness-prediction} further reports the Spearman correlation between recovery accessibility and maximum future SSCD. Since Immediate SSCD may influence both quantities, we also report partial Spearman correlation controlling for Immediate SSCD. We additionally report AUROC to evaluate how well recovery accessibility distinguishes reappearing targets from those that remain forgotten. Across methods, recovery accessibility consistently shows positive correlation with maximum future SSCD, and the association remains positive after controlling for Immediate SSCD. The AUROC values further indicate meaningful predictive ability for subsequent reappearance.

These findings show that immediate forgetting alone does not fully determine whether a target will remain forgotten. Even when the target satisfies the immediate SSCD criterion, nearby representations may remain easy for the model to denoise. This residual local recoverability is associated with later reappearance and motivates explicitly targeting the most recoverable local representation during deletion, forming the basis of \textsc{LASTING} in Section~\ref{sec:method}.

\section{Proposed Method}
\label{sec:method}
We propose \emph{Low-loss Adaptive Search and Targeted Intervention in
Neighborhood Geometry} (\textsc{LASTING}). At each optimization step,
\textsc{LASTING} finds a low-loss neighbor of the deletion target and uses it
to guide the base unlearning objective toward the model's strongest remaining
recovery route.

\subsection{Adaptive Low-Loss Neighbor Search}
\label{subsec:lasting_neighbor_search}

Let $\bar{\alpha}_t$ denote the cumulative diffusion noise schedule and let
$\xi=(t,\varepsilon)$ contain a timestep $t$ and Gaussian noise
$\varepsilon\sim\mathcal{N}(0,I)$.
For an image $a\in\mathcal{X}$, the noisy input and its standard DDPM
denoising loss~\citep{Ho_2020_NeurIPS} are
\begin{equation}
\begin{aligned}
    z_t(a;\xi)
    &=
    \sqrt{\bar{\alpha}_t}\,a
    +
    \sqrt{1-\bar{\alpha}_t}\,\varepsilon, \ \ \ 
    \ell_{\mathrm{diff}}(\theta;a,\xi)
    =
    \frac{1}{d}
    \left\|
        \varepsilon_{\theta}\left(z_t(a;\xi),t\right)
        -
        \varepsilon
    \right\|_2^2,
\end{aligned}
\label{eq:lasting_diffusion_loss}
\end{equation}
where $d$ is the dimensionality of the noise prediction.
A low value of $\ell_{\mathrm{diff}}$ indicates that the model can accurately
denoise the corresponding representation, suggesting that substantial local
recovery ability remains.

For target $a_i$, we define the feasible neighborhood
\begin{equation}
    \mathcal{D}_i
    =
    \left\{
        \delta:
        \|\delta\|_{\infty}\leq\epsilon_{\mathrm{nbr}},
        \;
        a_i+\delta\in\mathcal{X}
    \right\},
    \label{eq:lasting_neighborhood}
\end{equation}
where $\epsilon_{\mathrm{nbr}}$ controls the search radius and
$\mathcal{X}$ denotes the valid normalized image domain.
At iteration $q$, we sample $\xi_{i,q}$ and define
\begin{equation}
    f_{i,q}(\delta)
    =
    \ell_{\mathrm{diff}}
    \left(
        \theta_{i,q};
        a_i+\delta,
        \xi_{i,q}
    \right).
    \label{eq:lasting_inner_objective}
\end{equation}

Starting from $\delta_{i,q}^{(0)}=0$, we perform $J$ projected sign-gradient descent steps: \begin{equation} \begin{aligned} \delta_{i,q}^{(r+1)} = \Pi_{\mathcal{D}_i} \left[ \delta_{i,q}^{(r)} - \eta_{\mathrm{in}} \operatorname{sign} \left( \nabla_{\delta} f_{i,q} \left( \delta_{i,q}^{(r)} \right) \right) \right],  \\ 
\qquad r=0,\ldots,J-1. \end{aligned} \label{eq:lasting_neighbor_search} \end{equation} The resulting deletion representation is \begin{equation} \widetilde{a}_{i,q} = \operatorname{sg} \left( a_i+\delta_{i,q}^{(J)} \right). \label{eq:lasting_deletion_representation} \end{equation}
Here, $\eta_{\mathrm{in}}$ is the inner-search step size,
$\Pi_{\mathcal{D}_i}$ enforces the perturbation constraint and valid image
range, and $\operatorname{sg}$ denotes stop-gradient. We fix
$\xi_{i,q}$ during the inner search and reuse it for the subsequent deletion
update so that both steps use the same stochastic diffusion state.

\subsection{Recovery-Guided Unlearning}
\label{subsec:lasting_update}
Let $\mathcal{L}_{\mathrm{U}}$ denote a base unlearning objective.
\textsc{LASTING} replaces the original target $a_i$ with the selected neighbor
$\widetilde{a}_{i,q}$ while preserving the remaining components of the base
method:

\begin{equation}
    \mathcal{L}_{i,q}^{\mathrm{LASTING}}
    =
    \mathcal{L}_{\mathrm{U}}
    \left(
        \theta_{i,q};
        \mathcal{R}_{i,q},
        \widetilde{a}_{i,q},
        \xi_{i,q}
    \right).
    \label{eq:lasting_objective}
\end{equation}

Here, $\mathcal{R}_{i,q}$ is the retain mini-batch and $\xi_{i,q}$ is the
sampled diffusion state. The model is updated using the original optimization
rule, so \textsc{LASTING} changes only the representation used by the deletion
objective.

In our main implementation, we instantiate \textsc{LASTING} on
SISS~\citep{Alberti_2025_ICLR}. The retain construction, defensive mixture,
importance weighting, gradient normalization, clipping, and optimizer remain
unchanged. 
Let $g^R_{i,q}$ and $g^F_{i,q}$ denote the retain and forget
gradient components of SISS, respectively, with $g^F_{i,q}$ evaluated at
the selected low-loss neighbor $\widetilde{a}_{i,q}$. The resulting
\textsc{LASTING} gradient is

\begin{equation}
    g_{i,q}^{\mathrm{LASTING}}
    =
    g_{i,q}^{R}
    -
    \kappa
    \frac{
        g_{i,q}^{F}
    }{
        \left\|g_{i,q}^{F}\right\|_2
    }.
    \label{eq:lasting_siss_gradient}
\end{equation}

Then the model update is

\begin{equation}
    \theta_{i,q+1}
    =
    \operatorname{OptStep}_{\mathrm{SISS}}
    \left(
        \theta_{i,q};
        g_{i,q}^{\mathrm{LASTING}}
    \right).
    \label{eq:lasting_siss_update}
\end{equation}

Here, $\kappa$ is the target norm of the deletion-gradient component, and
$\operatorname{OptStep}_{\mathrm{SISS}}$ denotes the original SISS optimizer
step, including gradient clipping. 
Thus, \textsc{LASTING} modifies only the
representation used to compute the forget-side gradient, while preserving the
retain-side computation and the original SISS optimization procedure. This formulation allows \textsc{LASTING} to retain the original SISS update structure while incorporating the adaptively selected low-loss neighbor.

The selected neighbor $\widetilde{a}_{i,q}$ is detached before evaluating the
outer objective. Therefore, the outer update does not differentiate through
the neighbor search and requires no second-order derivatives. After $Q$
optimization iterations, we set $\theta_i=\theta_{i,Q}$ and proceed to the
next deletion request.

\section{Experiments}
\label{sec:experiments}
We now evaluate whether \textsc{LASTING} reduces recovery accessibility and improves
forgetting persistence under sequential deletion while maintaining generation quality on CelebA-HQ and Stable Diffusion.

\paragraph{Compared methods.}
We compare \textsc{LASTING} with SISS, EraseDiff, ReTrack, Forward
KL-Constrained Unlearning, and Prompt-Free Instance
Unlearning~\citep{Alberti_2025_ICLR,Wu_2025_CVPR,Shi_2026_AISTATS,Khalafi_2026_ICML,Lee_2026_CVPR_Workshop_MUV}.
The target order, retain set, model initialization, and evaluation protocol
are fixed across methods, while each baseline uses its recommended
optimization configuration and update budget. Prompt-Free additionally uses a generative model to
produce edited surrogate targets.

\subsection{Sequential Unlearning on CelebA-HQ}
\paragraph{Experimental setup.}
We use CelebA-HQ~\citep{Karras_2018_ICLR} and a pretrained
$256\times256$ unconditional DDPM~\citep{Ho_2020_NeurIPS}. We process
$K\in\{50,100,200,300\}$ deletion targets sequentially with the same model and
fix the target order and retain set across methods. Optimization details are
provided in the appendix.

\paragraph{Evaluation metrics.}
We report Immediate and Final SSCD~\citep{Pizzi_2022_CVPR}, Final Success and
Final Reappearance from Eq.~\ref{eq:persistence_metrics}, and
FID~\citep{Heusel_2017_NeurIPS}. Lower SSCD and FID indicate stronger
forgetting and better generation quality, respectively. We use the forgetting
threshold $\tau=0.7$.

\paragraph{Main comparison.}
Table~\ref{tab:celeba-sequential-results} compares methods with the deletion
target set and order fixed. 

At $K=300$, \textsc{LASTING} achieves $96.78\%$ Final Success and $3.12\%$
Final Reappearance, compared with $91.67\%$ and $7.93\%$ for SISS.
EraseDiff and ReTrack have Final Reappearance rates of $53.89\%$ and
$49.84\%$, respectively. 
\textsc{LASTING} also lowers Final SSCD relative to SISS ($0.4034$ versus
$0.4243$), with a FID of $23.14$. 

The advantages in Final Success and Final Reappearance extend across the evaluated sequence lengths and training seeds.
Mean $\pm$ sample standard deviation across training seeds is reported in Appendix~A.1. Figure~
\ref{fig:reappearance-count} shows how reappearance develops during the
reference deletion sequence. A more detailed trajectory-level analysis,
including transient and ever reappearance, is provided in Appendix~A.3.

\begin{table}[t]
\caption{Sequential data-point unlearning on CelebA-HQ for $K$ deletion
requests (\% of data). Immediate SSCD is measured after each deletion, and
final metrics at the last checkpoint. Final Success uses all $K$ targets,
while Final Reappearance uses immediately forgotten targets. Results are
means over three training seeds with a fixed target set and order.}
\label{tab:celeba-sequential-results}
\centering
\normalsize
\setlength{\tabcolsep}{2.5pt}
\renewcommand{\arraystretch}{1.0}

\begin{tabular}{@{}clccccc@{}}
\toprule
\shortstack{$K$\\(\% data)}
& Method
& \shortstack{Immediate\\SSCD $\downarrow$}
& \shortstack{Final\\SSCD $\downarrow$}
& \shortstack{Final\\Success (\%)$\uparrow$}
& \shortstack{Final\\Reapp. (\%)$\downarrow$}
& FID $\downarrow$ \\
\midrule

\multirow{6}{*}{\shortstack{$50$\\(0.17\%)}}
& SISS        & 0.3661 & 0.4704 & 98.00\% & 2.00\%  & 21.66 \\
& EraseDiff   & \textbf{0.1759} & \textbf{0.3522} & 86.67\% & 12.30\% & 73.76 \\
& ReTrack     & 0.3498 & 0.4599 & 92.00\% & 2.21\%  & 24.92 \\
& FU          & 0.2673 & 0.3575 & 99.33\% & 0.67\%  & 34.74 \\
& Prompt-Free & 0.4509 & 0.6159 & 77.33\% & 22.67\% & \textbf{19.71} \\
\cmidrule(lr){2-7}
\rowcolor{gray!18}
& \textsc{LASTING}
& 0.3087 & 0.4016 & \textbf{100.00\%} & \textbf{0.00\%} & 21.37 \\
\midrule

\multirow{6}{*}{\shortstack{$100$\\(0.33\%)}}
& SISS        & 0.3465 & 0.4530 & 96.33\% & 3.67\%  & \textbf{18.91} \\
& EraseDiff   & \textbf{0.1814} & 0.4684 & 72.67\% & 26.24\% & 152.08 \\
& ReTrack     & 0.3270 & 0.4481 & 89.00\% & 3.29\%  & 24.50 \\
& FU          & 0.2750 & \textbf{0.3972} & 96.67\% & 3.33\%  & 34.32 \\
& Prompt-Free & 0.4225 & 0.6124 & 75.33\% & 24.67\% & 19.59 \\
\cmidrule(lr){2-7}
\rowcolor{gray!18}
& \textsc{LASTING}
& 0.2906 & 0.3994 & \textbf{98.67\%} & \textbf{1.33\%} & 20.79 \\
\midrule

\multirow{6}{*}{\shortstack{$200$\\(0.67\%)}}
& SISS        & 0.3153 & 0.4244 & 95.67\% & 4.33\%  & 21.60 \\
& EraseDiff   & \textbf{0.1886} & 0.6356 & 47.33\% & 51.99\% & 168.25 \\
& ReTrack     & 0.3351 & 0.5254 & 80.33\% & 10.23\% & 21.76 \\
& FU          & 0.2560 & \textbf{0.3652} & 87.83\% & 12.06\% & 36.92 \\
& Prompt-Free & 0.3990 & 0.6554 & 55.83\% & 44.17\% & 20.66 \\
\cmidrule(lr){2-7}
\rowcolor{gray!18}
& \textsc{LASTING}
& 0.2651 & 0.3971 & \textbf{96.83\%} & \textbf{3.01\%} & \textbf{19.61} \\
\midrule

\multirow{6}{*}{\shortstack{$300$\\(1.00\%)}}
& SISS        & 0.2917 & 0.4243 & 91.67\% & 7.93\%  & \textbf{21.76} \\
& EraseDiff   & \textbf{0.1863} & 0.6322 & 45.44\% & 53.89\% & 134.95 \\
& ReTrack     & 0.3517 & 0.6739 & 43.11\% & 49.84\% & 28.30 \\
& FU          & 0.2240 & \textbf{0.3305} & 80.89\% & 19.06\% & 33.36 \\
& Prompt-Free & 0.3831 & 0.6751 & 47.67\% & 52.33\% & 21.91 \\
\cmidrule(lr){2-7}
\rowcolor{gray!18}
& \textsc{LASTING}
& 0.2365 & 0.4034 & \textbf{96.78\%} & \textbf{3.12\%} & 23.14 \\
\bottomrule
\end{tabular}
\end{table}

\paragraph{Different deletion target sets.}
The separate target-set comparison in
Table~\ref{tab:celeba-target-robustness} further shows that the improvement
persists when the deletion targets are changed at $K=200$.

\begin{table}[t]
\normalsize
\setlength{\belowcaptionskip}{\baselineskip}
\caption{Performance across three deletion configurations on CelebA-HQ
at $K=200$. $\mathcal{T}_1$, $\mathcal{T}_2$, and $\mathcal{T}_3$
start at indices $10{,}000$, $15{,}000$, and $20{,}000$.
Succ. denotes Final Success over all 200 targets, and Reapp. denotes
Final Reappearance among targets forgotten immediately after deletion.
The last column gives the mean and sample
SD of Reapp. across the three configurations.}
\label{tab:celeba-target-robustness}
\centering
\setlength{\tabcolsep}{5pt}
\renewcommand{\arraystretch}{1.12}

\begin{tabular}{lccccccc}
\toprule
& \multicolumn{2}{c}{$\mathcal{T}_1$ (Ref.)}
& \multicolumn{2}{c}{$\mathcal{T}_2$}
& \multicolumn{2}{c}{$\mathcal{T}_3$}
& Mean $\pm$ SD \\
Method
& Succ. & Reapp.
& Succ. & Reapp.
& Succ. & Reapp.
& Reapp. \\
\midrule
SISS
& 94.50\% & 5.50\%
& 79.50\% & 20.50\%
& 85.50\% & 14.50\%
& (13.50$\pm$7.55)\% \\
EraseDiff
& 63.00\% & 36.68\%
& 92.50\% & 7.22\%
& 58.00\% & 40.82\%
& (28.24$\pm$18.32)\% \\
ReTrack
& 84.50\% & 8.43\%
& 83.00\% & 12.83\%
& 68.50\% & 20.12\%
& (13.79$\pm$5.90)\% \\
FU
& 65.50\% & 34.17\%
& 98.50\% & 1.01\%
& 87.00\% & 13.00\%
& (16.06$\pm$16.79)\% \\
Prompt-Free
& 63.00\% & 37.00\%
& 58.50\% & 40.91\%
& 57.00\% & 43.00\%
& (40.30$\pm$3.05)\% \\
\midrule
\rowcolor{gray!18}
\textsc{LASTING}
& \textbf{97.50\%} & \textbf{2.50\%}
& \textbf{99.00\%} & \textbf{1.00\%}
& \textbf{99.00\%} & \textbf{1.00\%}
& $\boldsymbol{(1.50\pm0.87)\%}$ \\
\bottomrule
\end{tabular}
\end{table}

\subsection{Extension to Text-Conditioned Latent Diffusion}

\paragraph{Experimental setup and metrics.}
We further evaluate \textsc{LASTING} on Stable Diffusion v1.4, a
text-conditioned latent diffusion model~\citep{Rombach_2022_CVPR}.
We construct a sequence of $K_{\mathrm{SD}}=30$ memorized deletion targets
using the prompts identified by \citet{Webster_2023_arXiv}. Each target
consists of a reference memorized image and its original and modified
evaluation prompts. Following \citet{Alberti_2025_ICLR}, we generate 128
samples for each modified prompt to construct memorized and non-memorized
sets. At $K\in\{10,20,30\}$, we generate 16 images per target--prompt
pair with fixed random seeds and report Final Success, Reappearance Rate,
and CLIP-IQA~\citep{Wang_2023_AAAI}.
Further details are provided in the appendix.

\paragraph{Results.}
Table~\ref{tab:sd-sequential-results} shows that \textsc{LASTING}'s
persistence advantage extends to prompt-conditioned latent diffusion.
At $K=30$,
\textsc{LASTING} achieves 88.33\% Final Success with only 5.88\%
reappearance, compared with 66.67\%/22.45\% for SISS, 61.67\%/31.25\% for EraseDiff, and 93.33\%/6.90\% for ReTrack. 
These results indicate that \textsc{LASTING} preserves forgetting under
subsequent deletion updates. \textsc{LASTING} also maintains competitive
generation quality. The qualitative examples in Figure~\ref{fig:sd-qualitative} show that memorized content remains suppressed without obvious visual degradation in the shown examples.

\begin{table}[t]
\caption{{Sequential data-point unlearning on Stable Diffusion.}
Final Suc. denotes Final Success after $K$ deletion requests.
Among cases successfully forgotten immediately after deletion,
Reapp. measures reappearance under the final model.
IQA denotes CLIP-IQA generation quality.}
\label{tab:sd-sequential-results}
\centering
\setlength{\tabcolsep}{2.3pt}
\renewcommand{\arraystretch}{1.0}

\begin{tabular}{@{}l*{9}{c}@{}}
\toprule
& \multicolumn{3}{c}{$K=10$}
& \multicolumn{3}{c}{$K=20$}
& \multicolumn{3}{c}{$K=30$} \\
\cmidrule(lr){2-4}
\cmidrule(lr){5-7}
\cmidrule(lr){8-10}

Method
& \shortstack{Final Suc.\\$\uparrow$}
& \shortstack{Reapp.\\$\downarrow$}
& \shortstack{IQA\\$\uparrow$}
& \shortstack{Final Suc.\\$\uparrow$}
& \shortstack{Reapp.\\$\downarrow$}
& \shortstack{IQA\\$\uparrow$}
& \shortstack{Final Suc.\\$\uparrow$}
& \shortstack{Reapp.\\$\downarrow$}
& \shortstack{IQA\\$\uparrow$} \\
\midrule

SISS
& 90.00\% & \textbf{0.00\%} & 0.48
& 80.00\% & 12.12\% & 0.66
& 66.67\% & 22.45\% & 0.37 \\

EraseDiff
& 85.00\% & 14.29\% & 0.48
& 75.00\% & 18.75\% & 0.55
& 61.67\% & 31.25\% & 0.68 \\

ReTrack
& \textbf{100.00\%} & \textbf{0.00\%} & \textbf{0.71}
& 92.50\% & 7.50\% & \textbf{0.78}
& \textbf{93.33\%} & 6.90\% & 0.62 \\

FU
& 95.00\% & \textbf{0.00\%} & 0.58
& \textbf{97.50\%} & \textbf{2.70\%} & 0.74
& 61.67\% & 31.25\% & \textbf{0.84} \\

Prompt-Free
& 75.00\% & 11.76\% & 0.56
& 70.00\% & 15.63\% & 0.74
& 81.67\% & 10.20\% & 0.68 \\

\midrule
\rowcolor{gray!18}
\textsc{LASTING}
& \textbf{100.00\%} & \textbf{0.00\%} & 0.48
& 92.50\% & \textbf{2.70\%} & 0.62
& 88.33\% & \textbf{5.88\%} & 0.66 \\

\bottomrule
\end{tabular}
\end{table}

\begin{figure*}[t]
\centering

\newcommand{\sdqualcell}[2]{%
  \makebox[0.153\textwidth][c]{%
    \IfFileExists{#1}{%
      \includegraphics[width=0.153\textwidth]{#1}%
    }{%
      \fbox{%
        \parbox[c][0.136\textwidth][c]{0.136\textwidth}{%
          \centering #2%
        }%
      }%
    }%
  }%
}

\newcommand{\sdqualheader}[1]{%
  \makebox[0.153\textwidth][c]{%
    \footnotesize\bfseries #1%
  }%
}

\renewcommand{\arraystretch}{1.0}

\begin{tabular}{
@{}c
@{\hspace{5pt}}c
@{\hspace{5pt}}c
@{\hspace{5pt}}c
@{\hspace{5pt}}c
@{\hspace{5pt}}c@{}
}
&
\sdqualheader{Reference}
&
\sdqualheader{LASTING (Ours)}
&
\sdqualheader{SISS}
&
\sdqualheader{ReTrack}
&
\sdqualheader{EraseDiff}
\\[2pt]

\begin{tabular}{@{}c@{}}
  \parbox[c][0.153\textwidth][c]{0.118\textwidth}{%
    \centering\textbf{Immediate}\\\textbf{Model}%
  }\\[2pt]
  \parbox[c][0.153\textwidth][c]{0.118\textwidth}{%
    \centering\textbf{Final}\\\textbf{Model}%
  }
\end{tabular}
&
\parbox[c][0.31\textwidth][c]{0.153\textwidth}{%
  \centering
  \sdqualcell
    {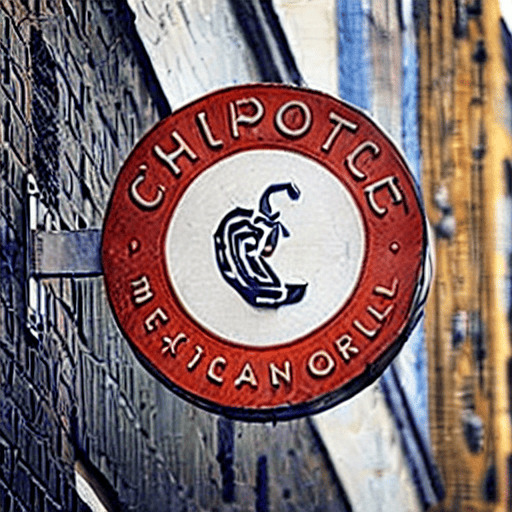}
    {Reference}%
}
&
\begin{tabular}{@{}c@{}}
  \sdqualcell
    {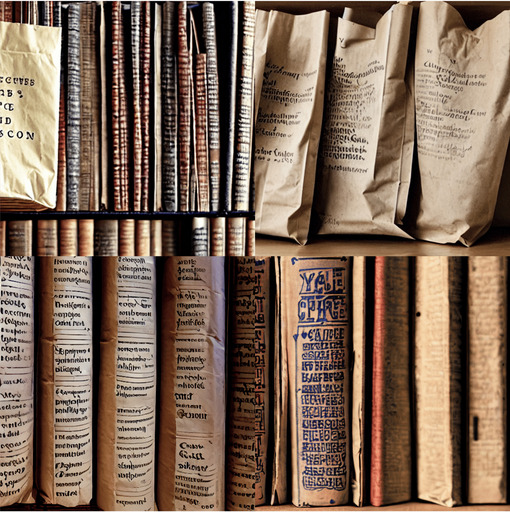}
    {LASTING\\Immediate}\\[2pt]
  \sdqualcell
    {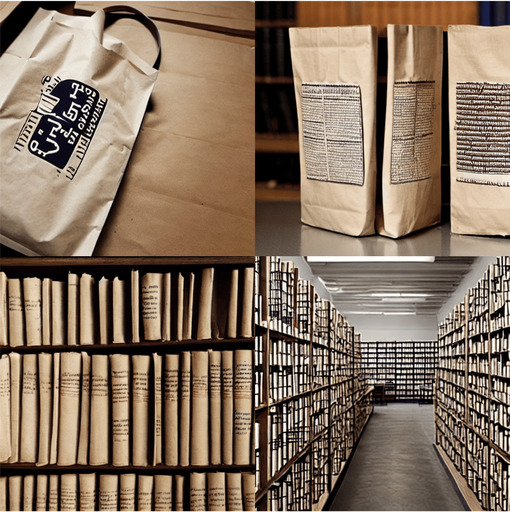}
    {LASTING\\Final}
\end{tabular}
&
\begin{tabular}{@{}c@{}}
  \sdqualcell
    {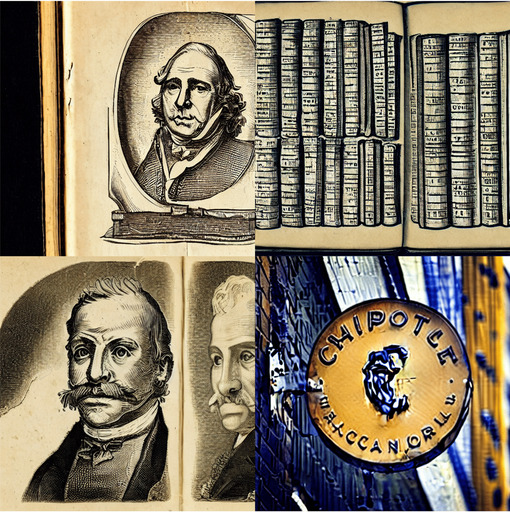}
    {SISS\\Immediate}\\[2pt]
  \sdqualcell
    {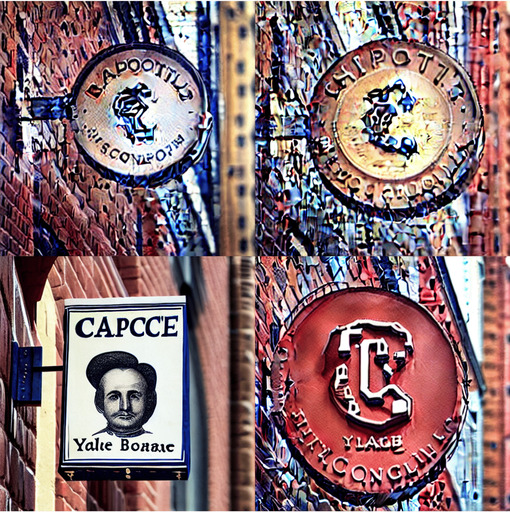}
    {SISS\\Final}
\end{tabular}
&
\begin{tabular}{@{}c@{}}
  \sdqualcell
    {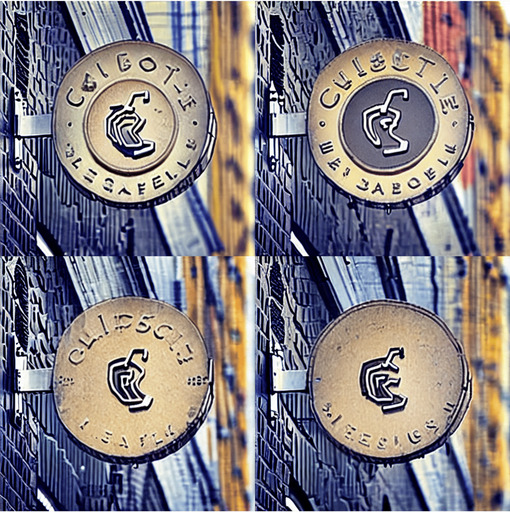}
    {ReTrack\\Immediate}\\[2pt]
  \sdqualcell
    {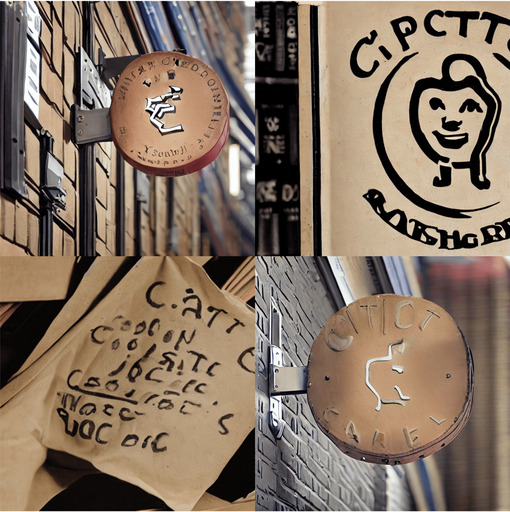}
    {ReTrack\\Final}
\end{tabular}
&
\begin{tabular}{@{}c@{}}
  \sdqualcell
    {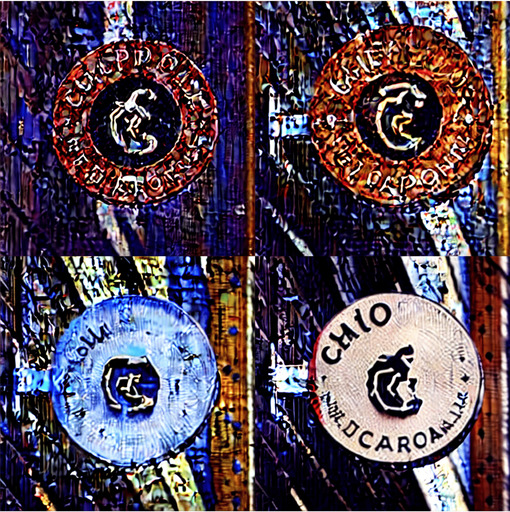}
    {EraseDiff\\Immediate}\\[2pt]
  \sdqualcell
    {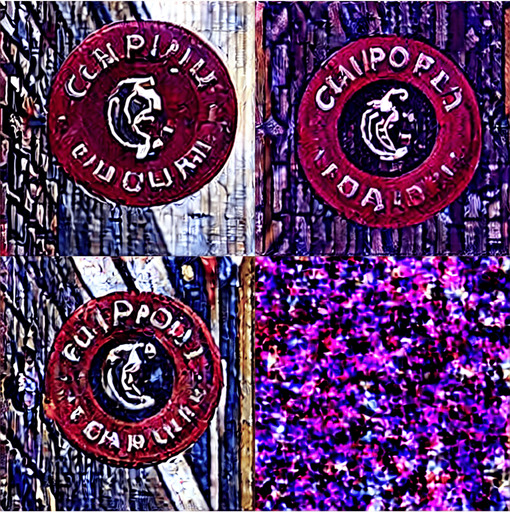}
    {EraseDiff\\Final}
\end{tabular}
\\
\end{tabular}

\caption{Qualitative comparison on Stable Diffusion.
The leftmost column shows the reference memorized image.
The remaining columns correspond to unlearning methods,
and rows show generations from the immediate and final model states.
Prompt: \textit{Chipotle Bag Essays Earn a Spot in Yale's Rare Book Library}}
\label{fig:sd-qualitative}
\end{figure*}

\subsection{Ablation and Analysis}
\paragraph{Neighbor construction.}
Table~\ref{tab:app_neighbor_ablation} shows that the direction of neighborhood
search is critical. \textsc{LASTING} achieves 2.50\% Final Reappearance and
97.50\% Final Success, compared with 5.50\% for SISS, 23.74\% for Random
Neighbor, and 37.20\% for High-Loss Neighbor. These results demonstrate that
explicitly suppressing a locally recoverable low-loss representation improves
persistent forgetting, whereas random perturbation and high-loss search degrade
performance. This contrast indicates that the benefit comes from searching toward low-loss, locally recoverable regions rather than from generic perturbation of the target neighborhood.

\begin{table}[t]
\caption{Effect of neighbor construction on sequential forgetting. Random
Neighbor uses a norm-matched random perturbation, High-Loss Neighbor performs
projected loss maximization, and \textsc{LASTING} performs projected loss
minimization. Final Success is measured over all 200 targets, whereas Final Reappearance is measured over targets forgotten immediately after deletion.}
\label{tab:app_neighbor_ablation}

\centering
\normalsize
\setlength{\tabcolsep}{3pt}
\renewcommand{\arraystretch}{1.0}

\begin{tabular}{@{}llccccc@{}}
\toprule
Variant
& \makecell{Neighbor\\construction}
& \makecell{Imm.\\SSCD $\downarrow$}
& \makecell{Final\\SSCD $\downarrow$}
& \makecell{Final\\Success $\uparrow$}
& \makecell{Final\\Reapp. $\downarrow$}
& FID $\downarrow$ \\
\midrule

\makecell[l]{SISS}
& \makecell[l]{Observed target\\$(\delta=0)$}
& \makecell{0.3024}
& \makecell{0.4286}
& \makecell{94.50\%}
& \makecell{5.50\%}
& \makecell{21.44} \\

\midrule

\makecell[l]{Random\\Neighbor}
& \makecell[l]{Norm-matched\\random perturbation}
& \makecell{0.4211}
& \makecell{0.5996}
& \makecell{75.50\%}
& \makecell{23.74\%}
& \makecell{\textbf{19.49}} \\

\midrule

\makecell[l]{High-Loss\\Neighbor}
& \makecell[l]{Projected loss\\maximization}
& \makecell{0.5082}
& \makecell{0.6517}
& \makecell{54.00\%}
& \makecell{37.20\%}
& \makecell{35.03} \\

\midrule

\makecell[l]{\textsc{LASTING}\\(Low-Loss Neighbor)}
& \makecell[l]{Projected loss\\minimization}
& \makecell{\textbf{0.2577}}
& \makecell{\textbf{0.3960}}
& \makecell{\textbf{97.50\%}}
& \makecell{\textbf{2.50\%}}
& \makecell{19.68} \\

\bottomrule
\end{tabular}
\end{table}

\paragraph{Recovery accessibility across target groups.}
Table~\ref{tab:app_recovery_sharpness_lasting} further shows that
\textsc{LASTING} reduces mean recovery accessibility from 0.0501 to 0.0309, a
38.35\% reduction. The reduction reaches 77.72\% and 82.83\% for the top 10\%
and 5\% most vulnerable targets, confirming that \textsc{LASTING} suppresses
the strongest remaining recovery geometry. Hyperparameter sensitivity and
computational overhead are reported in the appendix.

\newcommand{\reappsummary}[4]{%
  \shortstack{#1/#2 (#3\%)\\[-1pt]#4}%
}

\makeatletter
\newcommand{\sidetablecaption}[2]{%
  \def\@captype{table}%
  \caption{#2}%
  \label{#1}%
}
\makeatother

\begin{table}[t]
\centering

\begin{minipage}[t]{0.46\textwidth}
\centering
\sidetablecaption{tab:app_recovery_sharpness_lasting}{Recovery accessibility by \textsc{LASTING} across targets with different recovery accessibility levels. Reduction denotes the relative decrease
from SISS to \textsc{LASTING}.}

\normalsize
\setlength{\tabcolsep}{1pt}
\renewcommand{\arraystretch}{1.0}
\begin{tabular}{@{}lcccc@{}}
\toprule
& $N$ & SISS & \textsc{LASTING} & Reduction \\
\midrule
Entire set & 300 & 0.050 & 0.031 & 38.35\% \\
Top 25\%    & 75  & 0.161 & 0.051 & 68.39\% \\
Top 10\%    & 30  & 0.344 & 0.077 & 77.72\% \\
Top 5\%     & 15  & 0.558 & 0.096 & 82.83\% \\
\bottomrule
\end{tabular}
\end{minipage}\hfill
\begin{minipage}[t]{0.52\textwidth}
\centering
\sidetablecaption{tab:objective-compatibility}{Compatibility with different
unlearning objectives on CelebA-HQ at $K=200$. Final Reapp. denotes Final Reappearance among targets successfully forgotten immediately after deletion.}

\normalsize
\setlength{\tabcolsep}{3pt}
\renewcommand{\arraystretch}{1.0}
\begin{tabular}{@{}lc@{}}
\toprule
Method & Final Reapp. \\
\midrule
SISS & 5.50\% \\
\textsc{LASTING} + SISS & \textbf{2.50\%} \\
\midrule
FU & 34.17\% \\
\textsc{LASTING} + FU  & \textbf{4.50\%} \\
\bottomrule
\end{tabular}
\end{minipage}

\end{table}

\paragraph{Compatibility with different unlearning objectives.}
We further apply \textsc{LASTING} to Forward KL-Constrained Unlearning.
Table~\ref{tab:objective-compatibility} shows that it reduces Final
Reappearance from 5.50\% to 2.50\% for SISS and from 34.17\% to 4.50\% for FU.
The substantial improvement under FU indicates that recovery-guided target
selection remains effective across both evaluated unlearning objectives,
supporting the view that \textsc{LASTING}'s benefit arises from its
recovery-guided target modification rather than from the specific SISS objective.

\section{Conclusion}
\label{sec:conclusion}
In this paper, we identify \emph{sequential reappearance}, a failure mode in which a diffusion training instance that is forgotten immediately after deletion becomes recoverable again after subsequent unlearning updates. Our analysis indicates that this failure is associated with greater local recovery accessibility around the deleted target. Motivated by this finding, we propose \textsc{LASTING}, which adaptively identifies and suppresses locally recoverable representations during unlearning. Across diffusion settings, \textsc{LASTING} substantially improves forgetting persistence while maintaining competitive generation quality. These results highlight the need to evaluate diffusion unlearning beyond immediate deletion success and to account for the stability of forgetting under subsequent requests.

\subsection*{AI use statement}
In this work, we used generative AI tools for retrieval and discovery,
including identifying related work and suggesting search queries and keywords.
All retrieved sources were manually reviewed by the authors, who determined
the final selection and citation of related work.

\bibliography{iclr2027_conference}
\bibliographystyle{iclr2027_conference}

\appendix
\clearpage
\appendix

\section*{Supplementary Material Contents}

\noindent\textbf{A. Additional Experimental Analyses}\par
\hspace*{1.5em}A.1 Results Across Training Seeds\par
\hspace*{1.5em}A.2 Comparison at Matched Computation\par
\hspace*{1.5em}A.3 Detailed Analysis of Reappearance\par

\medskip
\noindent\textbf{B. Sensitivity Analyses}\par
\hspace*{1.5em}B.1 Number of Inner-Search Steps\par
\hspace*{1.5em}B.2 Neighborhood Radius\par
\hspace*{1.5em}B.3 Evaluation Timestep\par

\medskip
\noindent\textbf{C. Implementation Details}\par
\hspace*{1.5em}C.1 CelebA-HQ\par
\hspace*{1.5em}C.2 Stable Diffusion\par
\medskip

\noindent\textbf{D. Additional Qualitative Results}\par
\hspace*{1.5em}D.1 CelebA-HQ \par
\hspace*{1.5em}D.2 Stable Diffusion \par

\bigskip

\section{Additional Experimental Analyses}
\subsection{Results Across Training Seeds}
We repeat the CelebA-HQ sequential unlearning experiments with
training seeds 0, 42, and 10000 at $K\in\{50,100,200,300\}$.
For each sequence length, the deletion targets, their order, and the
evaluation protocol are fixed across seeds, and each method uses the
same optimization settings in all three runs.
Table~\ref{tab:celeba-sequential-results_2} reports the resulting
mean $\pm$ sample standard deviation.

\begin{table}[t]
\caption{Sequential data-point unlearning on CelebA-HQ for $K$ deletion
requests (\% of data). Immediate SSCD is measured after each deletion, and
final metrics at the last checkpoint. Final Success uses all $K$ targets,
while Final Reappearance uses immediately forgotten targets. Results are
reported as mean $\pm$ sample standard deviation over three training seeds
with a fixed target set and order. Standard deviations for percentage
metrics are expressed in percentage points.}
\label{tab:celeba-sequential-results_2}
\centering
\normalsize
\setlength{\tabcolsep}{1.5pt}
\renewcommand{\arraystretch}{1.0}

\resizebox{\textwidth}{!}{%
\begin{tabular}{@{}clccccc@{}}
\toprule
\shortstack{$K$\\(\% data)}
& Method
& \shortstack{Immediate\\SSCD $\downarrow$}
& \shortstack{Final\\SSCD $\downarrow$}
& \shortstack{Final\\Success (\%)$\uparrow$}
& \shortstack{Final\\Reapp. (\%)$\downarrow$}
& FID $\downarrow$ \\
\midrule

\multirow{6}{*}{\shortstack{$50$\\(0.17\%)}}
& SISS
& $0.3661 \pm 0.0102$
& $0.4704 \pm 0.0204$
& $98.00 \pm 2.00$
& $2.00 \pm 2.00$
& $21.66 \pm 3.11$ \\

& EraseDiff
& $\mathbf{0.1759 \pm 0.0125}$
& $\mathbf{0.3522 \pm 0.0895}$
& $86.67 \pm 10.07$
& $12.30 \pm 8.38$
& $73.76 \pm 29.75$ \\

& ReTrack
& $0.3498 \pm 0.0050$
& $0.4599 \pm 0.0178$
& $92.00 \pm 2.00$
& $2.21 \pm 2.13$
& $24.92 \pm 5.04$ \\

& FU
& $0.2673 \pm 0.0358$
& $0.3575 \pm 0.0096$
& $99.33 \pm 1.15$
& $0.67 \pm 1.15$
& $34.74 \pm 11.17$ \\

& Prompt-Free
& $0.4509 \pm 0.0538$
& $0.6159 \pm 0.0377$
& $77.33 \pm 11.02$
& $22.67 \pm 11.02$
& $\mathbf{19.71 \pm 2.00}$ \\

\cmidrule(lr){2-7}
\rowcolor{gray!18}
& \textsc{LASTING}
& $0.3087 \pm 0.0113$
& $0.4016 \pm 0.0117$
& $\mathbf{100.00 \pm 0.00}$
& $\mathbf{0.00 \pm 0.00}$
& $21.37 \pm 4.88$ \\
\midrule

\multirow{6}{*}{\shortstack{$100$\\(0.33\%)}}
& SISS
& $0.3465 \pm 0.0086$
& $0.4530 \pm 0.0385$
& $96.33 \pm 0.58$
& $3.67 \pm 0.58$
& $\mathbf{18.91 \pm 2.30}$ \\

& EraseDiff
& $\mathbf{0.1814 \pm 0.0185}$
& $0.4684 \pm 0.1092$
& $72.67 \pm 15.95$
& $26.24 \pm 15.05$
& $152.08 \pm 81.70$ \\

& ReTrack
& $0.3270 \pm 0.0107$
& $0.4481 \pm 0.0247$
& $89.00 \pm 1.73$
& $3.29 \pm 2.19$
& $24.50 \pm 3.49$ \\

& FU
& $0.2750 \pm 0.0257$
& $\mathbf{0.3972 \pm 0.0269}$
& $96.67 \pm 4.16$
& $3.33 \pm 4.16$
& $34.32 \pm 19.85$ \\

& Prompt-Free
& $0.4225 \pm 0.0417$
& $0.6124 \pm 0.0404$
& $75.33 \pm 11.72$
& $24.67 \pm 11.72$
& $19.59 \pm 2.14$ \\

\cmidrule(lr){2-7}
\rowcolor{gray!18}
& \textsc{LASTING}
& $0.2906 \pm 0.0113$
& $0.3994 \pm 0.0193$
& $\mathbf{98.67 \pm 0.58}$
& $\mathbf{1.33 \pm 0.58}$
& $20.79 \pm 3.16$ \\
\midrule

\multirow{6}{*}{\shortstack{$200$\\(0.67\%)}}
& SISS
& $0.3153 \pm 0.0113$
& $0.4244 \pm 0.0154$
& $95.67 \pm 1.04$
& $4.33 \pm 1.04$
& $21.60 \pm 0.66$ \\

& EraseDiff
& $\mathbf{0.1886 \pm 0.0212}$
& $0.6356 \pm 0.0337$
& $47.33 \pm 13.65$
& $51.99 \pm 13.36$
& $168.25 \pm 83.64$ \\

& ReTrack
& $0.3351 \pm 0.0038$
& $0.5254 \pm 0.0102$
& $80.33 \pm 4.01$
& $10.23 \pm 2.03$
& $21.76 \pm 5.98$ \\

& FU
& $0.2560 \pm 0.0064$
& $\mathbf{0.3652 \pm 0.2257}$
& $87.83 \pm 19.37$
& $12.06 \pm 19.17$
& $36.92 \pm 14.39$ \\

& Prompt-Free
& $0.3990 \pm 0.0352$
& $0.6554 \pm 0.0269$
& $55.83 \pm 6.79$
& $44.17 \pm 6.79$
& $20.66 \pm 0.76$ \\

\cmidrule(lr){2-7}
\rowcolor{gray!18}
& \textsc{LASTING}
& $0.2651 \pm 0.0081$
& $0.3971 \pm 0.0064$
& $\mathbf{96.83 \pm 0.58}$
& $\mathbf{3.01 \pm 0.50}$
& $\mathbf{19.61 \pm 0.49}$ \\
\midrule

\multirow{6}{*}{\shortstack{$300$\\(1.00\%)}}
& SISS
& $0.2917 \pm 0.0097$
& $0.4243 \pm 0.0043$
& $91.67 \pm 1.76$
& $7.93 \pm 1.60$
& $\mathbf{21.76 \pm 5.49}$ \\

& EraseDiff
& $\mathbf{0.1863 \pm 0.0181}$
& $0.6322 \pm 0.0195$
& $45.44 \pm 3.56$
& $53.89 \pm 3.78$
& $134.95 \pm 53.63$ \\

& ReTrack
& $0.3517 \pm 0.0082$
& $0.6739 \pm 0.0093$
& $43.11 \pm 3.67$
& $49.84 \pm 5.79$
& $28.30 \pm 4.47$ \\

& FU
& $0.2240 \pm 0.0245$
& $\mathbf{0.3305 \pm 0.2940}$
& $80.89 \pm 31.66$
& $19.06 \pm 31.58$
& $33.36 \pm 16.06$ \\

& Prompt-Free
& $0.3831 \pm 0.0317$
& $0.6751 \pm 0.0226$
& $47.67 \pm 6.49$
& $52.33 \pm 6.49$
& $21.91 \pm 1.96$ \\

\cmidrule(lr){2-7}
\rowcolor{gray!18}
& \textsc{LASTING}
& $0.2365 \pm 0.0089$
& $0.4034 \pm 0.0114$
& $\mathbf{96.78 \pm 1.02}$
& $\mathbf{3.12 \pm 1.08}$
& $23.14 \pm 5.41$ \\
\bottomrule
\end{tabular}%
}
\end{table}

\subsection{Comparison at Matched Computation}
\label{sec:app-computation}

\paragraph{Computational overhead.}
We compare the computational overhead of \textsc{LASTING} with SISS,
EraseDiff, ReTrack, Forward KL-Constrained Unlearning (FU), and
Prompt-Free Instance Unlearning. Each method uses its reported
CelebA-HQ configuration and update budget. We measure per-target
wall-clock time and peak process GPU memory on a single NVIDIA RTX
PRO 6000 Blackwell Server Edition GPU. As shown in
Table~\ref{tab:app-compute-overhead}, \textsc{LASTING} takes 960.43
seconds per target due to its inner search. Its peak GPU memory is
13.63 GiB, the lowest among the evaluated methods.

\begin{table}[ht!]
\caption{Computational overhead on CelebA-HQ. Each method uses its
reported optimization configuration and update budget.}
\label{tab:app-compute-overhead}
\centering
\normalsize
\setlength{\tabcolsep}{5pt}
\renewcommand{\arraystretch}{1.0}
\begin{tabular}{@{}lcc@{}}
\toprule
Method
& \shortstack{Time per target\\(s) $\downarrow$}
& \shortstack{Peak process GPU\\memory (GiB) $\downarrow$} \\
\midrule
SISS        & 257.70 & 13.83 \\
EraseDiff   & 252.92 & 21.14 \\
ReTrack     & 222.29 & 19.30 \\
FU          & 215.14 & 19.51 \\
Prompt-Free & 281.33 & 21.56 \\
\rowcolor{gray!18}
\textsc{LASTING} & 960.43 & 13.63 \\
\bottomrule
\end{tabular}
\end{table}
\paragraph{Increased-budget SISS control.}
We further evaluate whether additional SISS optimization reproduces
the persistence achieved by recovery-guided target replacement.
To approximately match the computational budget of \textsc{LASTING},
we increase the number of SISS training steps per deletion target based on
the measured per-target computation time of standard SISS and \textsc{LASTING}.
Table~\ref{tab:app-increased-budget} compares the reference SISS run,
the increased-budget SISS run, and \textsc{LASTING} along the deletion
sequence.

The increased-budget SISS run obtains the lowest Immediate
SSCD at every reported checkpoint, showing stronger forgetting
immediately after each deletion. Its Final Reappearance nevertheless
reaches 29.50\% at $K=200$ and 31.00\% at $K=300$, compared with
5.50\% and 9.73\% for reference SISS. \textsc{LASTING} reduces these
rates to 2.50\% and 2.33\%, respectively. At $K=300$, it also
maintains 97.67\% Final Success and a Final SSCD of 0.3917, compared
with 69.00\% and 0.5377 for increased-budget SISS. Increasing the
SISS optimization budget therefore strengthens immediate forgetting
without reproducing the persistence obtained by \textsc{LASTING}.

\begin{table}[ht!]
\caption{Sequential unlearning results for reference SISS,
increased-budget SISS, and \textsc{LASTING} on the recorded CelebA-HQ
deletion sequence. These are single-run results. Final Success is
measured over all $K$ targets. Final Reappearance is measured among
targets successfully forgotten immediately after deletion.}
\label{tab:app-increased-budget}
\centering
\normalsize
\setlength{\tabcolsep}{1.5pt}
\renewcommand{\arraystretch}{1.0}
\begin{tabular}{@{}clcccc@{}}
\toprule
$K$ & Method
& \shortstack{Immediate\\SSCD $\downarrow$}
& \shortstack{Final\\SSCD $\downarrow$}
& \shortstack{Final\\Success (\%) $\uparrow$}
& \shortstack{Final\\Reappearance $\downarrow$} \\
\midrule
\multirow{3}{*}{$50$}
& SISS                  & 0.3545 & 0.4739 & 100.00 & 0/50 (0.00\%) \\
& SISS (extended)       & 0.1539 & 0.3819 & 96.00  & 2/50 (4.00\%) \\
\rowcolor{gray!18}
& \textsc{LASTING}      & 0.2965 & 0.3944 & 100.00 & 0/50 (0.00\%) \\
\midrule
\multirow{3}{*}{$100$}
& SISS                  & 0.3366 & 0.4171 & 96.00 & 4/100 (4.00\%) \\
& SISS (extended)       & 0.1161 & 0.5279 & 89.00 & 11/100 (11.00\%) \\
\rowcolor{gray!18}
& \textsc{LASTING}      & 0.2807 & 0.3785 & 99.00 & 1/100 (1.00\%) \\
\midrule
\multirow{3}{*}{$200$}
& SISS                  & 0.3024 & 0.4286 & 94.50 & 11/200 (5.50\%) \\
& SISS (extended)       & 0.0851 & 0.5691 & 70.50 & 59/200 (29.50\%) \\
\rowcolor{gray!18}
& \textsc{LASTING}      & 0.2577 & 0.3960 & 97.50 & 5/200 (2.50\%) \\
\midrule
\multirow{3}{*}{$300$}
& SISS                  & 0.2807 & 0.4238 & 89.67 & 29/298 (9.73\%) \\
& SISS (extended)       & 0.0664 & 0.5377 & 69.00 & 93/300 (31.00\%) \\
\rowcolor{gray!18}
& \textsc{LASTING}      & 0.2281 & 0.3917 & 97.67 & 7/300 (2.33\%) \\
\bottomrule
\end{tabular}
\end{table}

\subsection{Detailed Analysis of Reappearance}
\label{sec:app-trajectory-reappearance}

We further analyze reappearance over deletion trajectories of
$K\in\{50,100,200,300\}$.
For each target satisfying $s_i(\theta_i)<\tau$ immediately after
deletion, we examine all subsequent checkpoints using the same
SSCD threshold $\tau=0.7$ as in the main paper.
\emph{No Reappearance} indicates that the target remains below
the threshold at all subsequent checkpoints, whereas
\emph{Transient Reappearance} indicates that it reaches or exceeds
the threshold but returns below it at the final checkpoint.
\emph{Final Reappearance} indicates that the target is at or above
the threshold under the final model.
Ever Reappearance consists of transient and final reappearance.

\begin{table}[p]
\caption{Trajectory-level reappearance on CelebA-HQ for training
seed 42. Immediate Success is reported over all $K$ targets.
The remaining categories are reported over targets successfully
forgotten immediately after deletion. Best results for each $K$
are shown in bold.}
\label{tab:app-trajectory-reappearance}
\centering
\normalsize
\setlength{\tabcolsep}{3pt}
\renewcommand{\arraystretch}{1.0}

\newcommand{\trajcell}[2]{%
  \begin{tabular}[c]{@{}c@{}}#1\\(#2\%)\end{tabular}%
}

\begin{tabular}{@{}clccccc@{}}
\toprule
$K$ & Method
& \shortstack{Immediate\\Success $\uparrow$}
& \shortstack{No\\Reapp. $\uparrow$}
& \shortstack{Transient\\Reapp. $\downarrow$}
& \shortstack{Final\\Reapp. $\downarrow$}
& \shortstack{Ever\\Reapp. $\downarrow$} \\
\midrule

& SISS
& {\bfseries \trajcell{50/50}{100.00}}
& \trajcell{49/50}{98.00}
& \trajcell{1/50}{2.00}
& {\bfseries \trajcell{0/50}{0.00}}
& \trajcell{1/50}{2.00} \\
& EraseDiff
& {\bfseries \trajcell{50/50}{100.00}}
& \trajcell{33/50}{66.00}
& \trajcell{11/50}{22.00}
& \trajcell{6/50}{12.00}
& \trajcell{17/50}{34.00} \\
& ReTrack
& \trajcell{47/50}{94.00}
& \trajcell{45/47}{95.74}
& {\bfseries \trajcell{0/47}{0.00}}
& \trajcell{2/47}{4.26}
& \trajcell{2/47}{4.26} \\
& FU
& {\bfseries \trajcell{50/50}{100.00}}
& \trajcell{48/50}{96.00}
& \trajcell{2/50}{4.00}
& {\bfseries \trajcell{0/50}{0.00}}
& \trajcell{2/50}{4.00} \\
& Prompt-Free
& {\bfseries \trajcell{50/50}{100.00}}
& \trajcell{43/50}{86.00}
& \trajcell{2/50}{4.00}
& \trajcell{5/50}{10.00}
& \trajcell{7/50}{14.00} \\
\cmidrule(lr){2-7}
\rowcolor{gray!18}
\multirow{-12}{*}{$50$}
& \textsc{LASTING}
& {\bfseries \trajcell{50/50}{100.00}}
& {\bfseries \trajcell{50/50}{100.00}}
& {\bfseries \trajcell{0/50}{0.00}}
& {\bfseries \trajcell{0/50}{0.00}}
& {\bfseries \trajcell{0/50}{0.00}} \\
\midrule

& SISS
& {\bfseries \trajcell{100/100}{100.00}}
& \trajcell{91/100}{91.00}
& \trajcell{5/100}{5.00}
& \trajcell{4/100}{4.00}
& \trajcell{9/100}{9.00} \\
& EraseDiff
& {\bfseries \trajcell{100/100}{100.00}}
& \trajcell{59/100}{59.00}
& \trajcell{32/100}{32.00}
& \trajcell{9/100}{9.00}
& \trajcell{41/100}{41.00} \\
& ReTrack
& \trajcell{92/100}{92.00}
& \trajcell{87/92}{94.57}
& {\bfseries \trajcell{2/92}{2.17}}
& \trajcell{3/92}{3.26}
& \trajcell{5/92}{5.43} \\
& FU
& {\bfseries \trajcell{100/100}{100.00}}
& \trajcell{82/100}{82.00}
& \trajcell{10/100}{10.00}
& \trajcell{8/100}{8.00}
& \trajcell{18/100}{18.00} \\
& Prompt-Free
& {\bfseries \trajcell{100/100}{100.00}}
& \trajcell{60/100}{60.00}
& \trajcell{24/100}{24.00}
& \trajcell{16/100}{16.00}
& \trajcell{40/100}{40.00} \\
\cmidrule(lr){2-7}
\rowcolor{gray!18}
\multirow{-12}{*}{$100$}
& \textsc{LASTING}
& {\bfseries \trajcell{100/100}{100.00}}
& {\bfseries \trajcell{96/100}{96.00}}
& \trajcell{3/100}{3.00}
& {\bfseries \trajcell{1/100}{1.00}}
& {\bfseries \trajcell{4/100}{4.00}} \\
\midrule

& SISS
& {\bfseries \trajcell{200/200}{100.00}}
& \trajcell{173/200}{86.50}
& \trajcell{16/200}{8.00}
& \trajcell{11/200}{5.50}
& \trajcell{27/200}{13.50} \\
& EraseDiff
& \trajcell{199/200}{99.50}
& \trajcell{78/199}{39.20}
& \trajcell{48/199}{24.12}
& \trajcell{73/199}{36.68}
& \trajcell{121/199}{60.80} \\
& ReTrack
& \trajcell{178/200}{89.00}
& \trajcell{126/178}{70.79}
& \trajcell{37/178}{20.79}
& \trajcell{15/178}{8.43}
& \trajcell{52/178}{29.21} \\
& FU
& \trajcell{199/200}{99.50}
& \trajcell{72/199}{36.18}
& \trajcell{59/199}{29.65}
& \trajcell{68/199}{34.17}
& \trajcell{127/199}{63.82} \\
& Prompt-Free
& {\bfseries \trajcell{200/200}{100.00}}
& \trajcell{80/200}{40.00}
& \trajcell{46/200}{23.00}
& \trajcell{74/200}{37.00}
& \trajcell{120/200}{60.00} \\
\cmidrule(lr){2-7}
\rowcolor{gray!18}
\multirow{-11}{*}{$200$}
& \textsc{LASTING}
& {\bfseries \trajcell{200/200}{100.00}}
& {\bfseries \trajcell{191/200}{95.50}}
& {\bfseries \trajcell{4/200}{2.00}}
& {\bfseries \trajcell{5/200}{2.50}}
& {\bfseries \trajcell{9/200}{4.50}} \\
\midrule

& SISS
& \trajcell{298/300}{99.33}
& \trajcell{228/298}{76.51}
& \trajcell{41/298}{13.76}
& \trajcell{29/298}{9.73}
& \trajcell{70/298}{23.49} \\
& EraseDiff
& \trajcell{297/300}{99.00}
& \trajcell{57/297}{19.19}
& \trajcell{67/297}{22.56}
& \trajcell{173/297}{58.25}
& \trajcell{240/297}{80.81} \\
& ReTrack
& \trajcell{263/300}{87.67}
& \trajcell{92/263}{34.98}
& \trajcell{23/263}{8.75}
& \trajcell{148/263}{56.27}
& \trajcell{171/263}{65.02} \\
& FU
& \trajcell{299/300}{99.67}
& \trajcell{88/299}{29.43}
& \trajcell{45/299}{15.05}
& \trajcell{166/299}{55.52}
& \trajcell{211/299}{70.57} \\
& Prompt-Free
& {\bfseries \trajcell{300/300}{100.00}}
& \trajcell{94/300}{31.33}
& \trajcell{71/300}{23.67}
& \trajcell{135/300}{45.00}
& \trajcell{206/300}{68.67} \\
\cmidrule(lr){2-7}
\rowcolor{gray!18}
\multirow{-12}{*}{$300$}
& \textsc{LASTING}
& {\bfseries \trajcell{300/300}{100.00}}
& {\bfseries \trajcell{279/300}{93.00}}
& {\bfseries \trajcell{14/300}{4.67}}
& {\bfseries \trajcell{7/300}{2.33}}
& {\bfseries \trajcell{21/300}{7.00}} \\
\bottomrule
\end{tabular}
\end{table}

As shown in Table~\ref{tab:app-trajectory-reappearance},
\textsc{LASTING} achieves the lowest Ever Reappearance rate
across all four sequence lengths.
At $K=50$, none of its immediately forgotten targets reappear.
At $K=100$ and $K=200$, its Ever Reappearance rates are
$4.00\%$ and $4.50\%$, compared with $9.00\%$ and $13.50\%$
for SISS.
At $K=300$, \textsc{LASTING} maintains the highest
No Reappearance rate of $93.00\%$ and the lowest
Ever Reappearance rate of $7.00\%$.
Its $21$ reappearing targets consist of $14$ transient and
$7$ final cases.
In comparison, SISS exhibits an Ever Reappearance rate of
$23.49\%$, while the other baselines range from
$65.02\%$ to $80.81\%$.
These results show that \textsc{LASTING} reduces both transient
and final reappearance throughout the deletion trajectory. 

Subsequent deletion updates can also correct some immediate forgetting failures. Our reappearance metrics instead focus on cases where a successfully forgotten target later becomes memorized again, because this represents the loss of an already achieved forgetting state.

\section{Sensitivity Analyses}
\label{sec:app-sensitivity}

We conduct ablations on the number of inner-search steps $J$ and the
neighborhood radius $\epsilon_{\mathrm{nbr}}$ in \textsc{LASTING}.
All hyperparameter ablations use the CelebA-HQ sequential deletion
setting with SISS as the outer unlearning objective. Within each
ablation, we vary only the component under study while keeping the
deletion order, retain set, outer-update budget, optimizer, and
evaluation protocol fixed.

We follow the persistence metrics defined in the main paper using SSCD
similarity with the operational threshold $\tau=0.7$. Final Success
is normalized by all $K$ deletion targets, whereas Final Reappearance
is normalized by the immediate-success cohort. We report each rate as
both a raw count and a percentage. Immediate and Final SSCD are
averaged over all $K$ targets.


\subsection{Number of Inner-Search Steps}
We examine how the inner-search budget affects forgetting persistence
by varying $J\in\{1,4,8,12\}$ on the same $K=300$ deletion sequence.
All configurations use $\epsilon_{\mathrm{nbr}}=4/255$ and share the
same outer optimization and evaluation settings. The diffusion
timestep and noise are fixed throughout each inner search and reused
in the subsequent outer update.

\begin{table}[ht!]
\normalsize
\setlength{\belowcaptionskip}{\baselineskip}
\caption{Sensitivity to the number of inner-search steps.}
\label{tab:app-inner-steps}
\centering
\setlength{\tabcolsep}{6pt}
\renewcommand{\arraystretch}{1.08}
\begin{tabular}{@{}cccccc@{}}
\toprule
$J$
& \shortstack{Time per Target\\(s) $\downarrow$}
& \shortstack{Immediate\\SSCD $\downarrow$}
& \shortstack{Final\\SSCD $\downarrow$}
& \shortstack{Final\\Success $\uparrow$}
& \shortstack{Final\\Reapp. $\downarrow$} \\
\midrule
1 & 419.31 & 0.3142 & 0.5141
  & \shortstack{193/300\\(64.33\%)}
  & \shortstack{90/283\\(31.80\%)} \\
\addlinespace[2pt]
4 & 647.26 & 0.2382 & 0.4205
  & \shortstack{287/300\\(95.67\%)}
  & \shortstack{13/300\\(4.33\%)} \\
\addlinespace[2pt]
8 & 960.43 & \textbf{0.2281} & \textbf{0.3917}
  & \shortstack{\textbf{293/300}\\\textbf{(97.67\%)}}
  & \shortstack{\textbf{7/300}\\\textbf{(2.33\%)}} \\
\addlinespace[2pt]
12 & 1301.56 & 0.2932 & 0.4580
   & \shortstack{284/300\\(94.67\%)}
   & \shortstack{13/297\\(4.38\%)} \\
\bottomrule
\end{tabular}
\end{table}

As shown in Table~\ref{tab:app-inner-steps}, $J=8$ achieves the
lowest Final Reappearance rate of 2.33\% and the lowest Final SSCD
of 0.3917. Increasing the search budget from $J=4$ to $J=8$
improves both metrics, whereas $J=12$ provides no further benefit
and increases Final SSCD and endpoint reappearance. We therefore
use $J=8$ for the CelebA-HQ experiments.

\subsection{Neighborhood Radius}
Neighborhood radii are reported in the conventional $[0,1]$ pixel
scale. Since the implementation normalizes images to $[-1,1]$, a
radius of $r/255$ corresponds to $2r/255$ in the implementation
space. All perturbations are projected onto the corresponding
$\ell_\infty$ neighborhood and the valid normalized image domain.

We examine how the neighborhood size affects forgetting persistence
by varying $\epsilon_{\mathrm{nbr}}\in
\{4/255,8/255,12/255,16/255\}$ on the same $K=100$ deletion
sequence. All configurations use $J=4$ inner-search steps and share
the same outer optimization and evaluation settings. The reported
radii follow the conventional $[0,1]$ pixel scale.

\begin{table}[ht!]
\normalsize
\setlength{\belowcaptionskip}{\baselineskip}
\caption{Sensitivity to the neighborhood radius.}
\label{tab:app-neighborhood-radius}
\centering
\setlength{\tabcolsep}{6pt}
\renewcommand{\arraystretch}{1.08}
\begin{tabular}{@{}ccccc@{}}
\toprule
$\epsilon_{\mathrm{nbr}}$
& \shortstack{Immediate\\SSCD $\downarrow$}
& \shortstack{Final\\SSCD $\downarrow$}
& \shortstack{Final\\Success $\uparrow$}
& \shortstack{Final\\Reapp. $\downarrow$} \\
\midrule
$4/255$ & 0.2969 & 0.4000
& \shortstack{99/100\\(99.00\%)}
& \shortstack{0/99\\(0.00\%)} \\
\addlinespace[2pt]
$8/255$ & \textbf{0.2746} & \textbf{0.3495}
& \shortstack{\textbf{100/100}\\\textbf{(100.00\%)}}
& \shortstack{\textbf{0/100}\\\textbf{(0.00\%)}} \\
\addlinespace[2pt]
$12/255$ & 0.3969 & 0.5300
& \shortstack{81/100\\(81.00\%)}
& \shortstack{3/84\\(3.57\%)} \\
\addlinespace[2pt]
$16/255$ & 0.5626 & 0.7030
& \shortstack{44/100\\(44.00\%)}
& \shortstack{19/63\\(30.16\%)} \\
\bottomrule
\end{tabular}
\end{table}

As shown in Table~\ref{tab:app-neighborhood-radius},
$\epsilon_{\mathrm{nbr}}=8/255$ exhibits no final reappearance
and achieves a Final SSCD of 0.3495. Increasing the radius to
$12/255$ and $16/255$ raises Final SSCD to 0.5300 and 0.7030,
respectively, while also producing more final reappearance. These
results indicate that expanding the search region beyond $8/255$
weakens forgetting persistence under the tested setting. Although
$8/255$ performs best in this ablation, we use $4/255$ as a
conservative default for the longer main experiments.

\subsection{Evaluation Timestep}
We examine the stability of the evaluation results across two
evaluation timesteps, $t_{\mathrm{eval}}\in\{250,350\}$, on the \(K=300\) deletion sequence. For each timestep, results are aggregated over three recovery-noise seeds.

\begin{table}[ht!]
\normalsize
\setlength{\belowcaptionskip}{\baselineskip}
\caption{Sensitivity to evaluation timestep at $K=300$. For each evaluation
timestep, results are reported as mean $\pm$ sample standard
deviation over three recovery-noise seeds. The best result for each evaluation
timestep is shown in bold.}
\label{tab:app-evaluation-timestep}
\centering
\setlength{\tabcolsep}{6pt}
\renewcommand{\arraystretch}{1.08}
\begin{tabular}{@{}lccccc@{}}
\toprule
Method
& $t_{\mathrm{eval}}$
& \shortstack{Immediate\\SSCD $\downarrow$}
& \shortstack{Final\\SSCD $\downarrow$}
& \shortstack{Final\\Success (\%) $\uparrow$}
& \shortstack{Final\\Reapp. (\%) $\downarrow$} \\
\midrule
\multirow{2}{*}{SISS}
& 250 & 0.2831$\pm$0.0030 & 0.4244$\pm$0.0032
& 90.33$\pm$0.67 & 9.16$\pm$0.53 \\
& 350 & 0.2401$\pm$0.0026 & \textbf{0.4368$\pm$0.0024}
& 93.44$\pm$0.96 & 6.35$\pm$1.14 \\
\midrule
\multirow{2}{*}{EraseDiff}
& 250 & \textbf{0.1791$\pm$0.0099} & 0.6602$\pm$0.0053
& 40.00$\pm$1.33 & 59.78$\pm$1.46 \\
& 350 & \textbf{0.0861$\pm$0.0077} & 0.5670$\pm$0.0080
& 82.89$\pm$1.92 & 17.11$\pm$1.92 \\
\midrule
\multirow{2}{*}{ReTrack}
& 250 & 0.3429$\pm$0.0015 & 0.6858$\pm$0.0021
& 39.22$\pm$1.17 & 56.60$\pm$1.12 \\
& 350 & 0.3284$\pm$0.0030 & 0.6343$\pm$0.0009
& 64.44$\pm$0.69 & 31.16$\pm$1.28 \\
\midrule
\multirow{2}{*}{FU}
& 250 & 0.2320$\pm$0.0034 & 0.6175$\pm$0.0040
& 62.78$\pm$1.35 & 37.08$\pm$1.26 \\
& 350 & 0.2088$\pm$0.0045 & 0.5341$\pm$0.0037
& 94.22$\pm$1.39 & 5.78$\pm$1.39 \\
\midrule
\multirow{2}{*}{Prompt-Free}
& 250 & 0.3463$\pm$0.0006 & 0.6516$\pm$0.0026
& 55.11$\pm$0.84 & 44.89$\pm$0.84 \\
& 350 & 0.3426$\pm$0.0012 & 0.6249$\pm$0.0004
& 66.22$\pm$1.02 & 33.78$\pm$1.02 \\
\midrule
\multirow{2}{*}{\textsc{LASTING}}
& 250 & 0.2299$\pm$0.0036 & \textbf{0.3900$\pm$0.0015}
& \textbf{97.11$\pm$0.51} & \textbf{2.89$\pm$0.51} \\
& 350 & 0.1920$\pm$0.0012 & 0.4372$\pm$0.0029
& \textbf{97.44$\pm$0.51} & \textbf{2.56$\pm$0.51} \\
\bottomrule
\end{tabular}
\end{table}

As shown in Table~\ref{tab:app-evaluation-timestep},
\textsc{LASTING} maintains its forgetting-persistence advantage
across recovery-noise seeds and evaluation timesteps. This consistency
indicates that the reduction in sequential reappearance achieved
by \textsc{LASTING} persists across the evaluated recovery-noise seeds and evaluation settings, while the small standard deviations of
\textsc{LASTING} further demonstrate that its performance is
stable across the evaluated seeds.

\section{Implementation Details}
\label{sec:app-protocols}

We provide implementation and evaluation details for the CelebA-HQ and
Stable Diffusion experiments.

\subsection{CelebA-HQ}
\label{sec:app-celeba-protocol}

\paragraph{Dataset and sequential deletion protocol.}
We use the 30,000-image CelebA-HQ dataset and the pretrained
\texttt{google/ddpm-celebahq-256} unconditional DDPM. The $256\times256$
images are normalized to $[-1,1]$. The deletion sequence contains 300
targets, ordered from \texttt{10000.jpg} to \texttt{10299.jpg}, and we
report the prefix checkpoints at $K\in\{50,100,200,300\}$. After
unlearning target $a_i$, the resulting model $\theta_i$ initializes
deletion stage $i+1$.

The retain pool is fixed throughout the trajectory. All 300 deletion
targets are excluded before sequential unlearning begins, leaving
29,700 retain images. The target order, retain pool, and model
initialization are shared across all methods.

For recovery evaluation, each target is noised at diffusion timestep
$t=250$ and reconstructed through the reverse process from $t=250$
to $0$. We use one reconstruction per target and the same evaluation
seed and noise realization across all methods, including Prompt-Free
Instance Unlearning. Each reconstruction is saved as a PNG and evaluated
using the same SSCD encoder and ImageNet preprocessing. We use the
memorization threshold $\tau=0.7$. Immediate SSCD is measured using
$\theta_i$, whereas Final SSCD for a prefix of length $K$ is measured
using $\theta_K$.

FID is evaluated at each reported prefix checkpoint using 10,000
generated images, 50 DDPM sampling steps, and the full CelebA-HQ
reference set.

\paragraph{Compared methods and optimization details.}
We compare \textsc{LASTING} with SISS, EraseDiff, ReTrack, Forward
KL-Constrained Unlearning (FU), and Prompt-Free Instance Unlearning.
Each baseline uses the best-performing or recommended optimization
configuration from its original work. Table~\ref{tab:app-celeba-optimization}
reports the configurations used for the CelebA-HQ experiments.

\begin{table}[ht!]
\caption{Optimization settings for the CelebA-HQ experiments.
}
\label{tab:app-celeba-optimization}
\centering
\footnotesize
\setlength{\tabcolsep}{3pt}
\renewcommand{\arraystretch}{1.08}
\begin{tabular}{@{}lcccp{2.0in}@{}}
\toprule
Method & Optimizer & Learning rate & Batch
& Method-specific setting \\
\midrule
SISS
& AdamW & $5\times10^{-6}$ & 4
& $\lambda=0.5$, \texttt{scaling\_norm}$=500$ \\
EraseDiff
& AdamW & $5\times10^{-6}$ & 4
& $\eta=10^{-3}$ \\
ReTrack
& AdamW & $5\times10^{-6}$ & 4
& $k=10$, $\lambda=0.005$ \\
FU
& Adam & $5\times10^{-6}$ & 4
& $\epsilon_{\mathrm{KL}}=0.01$, $\eta_{\lambda}=0.2$ \\
Prompt-Free
& AdamW & $5\times10^{-6}$ & 4
& $\beta=5\times10^{-5}$, gradient surgery \\
\midrule
\rowcolor{gray!18}
\textsc{LASTING}
& AdamW & $5\times10^{-6}$ & 4
& $\epsilon_{\mathrm{nbr}}=4/255$, $J=8$ \\
\bottomrule
\end{tabular}
\end{table}

All methods use a constant learning-rate schedule without warmup and
are trained in FP32. SISS, EraseDiff, ReTrack, and \textsc{LASTING}
use a maximum gradient norm of $1.0$. FU uses $0.3$, and Prompt-Free
does not apply gradient clipping. Prompt-Free uses a fixed edited
surrogate for each target and applies its original gradient-surgery
update.

\paragraph{\textsc{LASTING} implementation.}
\textsc{LASTING} is implemented on top of SISS while leaving its
retain construction and outer deletion objective unchanged. The inner
search uses the raw DDPM $\epsilon$-prediction MSE. We set
$\epsilon_{\mathrm{nbr}}=4/255$, $J=8$, and
$\eta_{\mathrm{in}}=\epsilon_{\mathrm{nbr}}/J=0.5/255$ in the
conventional $[0,1]$ pixel scale. These values correspond to $8/255$
and $1/255$, respectively, in the implementation's $[-1,1]$ scale.

Each search starts from zero perturbation and performs projected
sign-gradient descent without random initialization. After each inner
step, the perturbation is projected onto the $\ell_\infty$
neighborhood, and the resulting image is clipped to $[-1,1]$. The
diffusion timestep and noise are fixed across the $J$ inner steps and
reused in the subsequent outer update. 

\paragraph{FU+LASTING implementation.}
We also apply the low-loss neighbor replacement of \textsc{LASTING}
to FU. The selected neighbor replaces the observed deletion target
in FU's forget-side computation. The retain construction and FU
optimization rule remain unchanged. FU and FU+LASTING use the same
deletion sequence and evaluation protocol.

\paragraph{Recovery accessibility measurement.}
We measure recovery accessibility at the checkpoint $\theta_i^{(u)}$
immediately after method $u$ deletes target $a_i$. For each target,
the fixed bank $\mathcal{B}_i$ contains timesteps
$\{50,100,150,200,250\}$ and two Gaussian noise draws per timestep.
We use the base seed $20260713$ with a target-specific index offset.
For each target, the same bank is reused across methods and for both
the observed target and all perturbed candidates evaluated during
the search.

We set $\epsilon_{\mathrm{acc}}=0.03137255\simeq8/255$ in the
normalized $[-1,1]$ image space, corresponding to $4/255$ in the
$[0,1]$ pixel scale. Starting from zero perturbation, we perform
eight projected sign-gradient descent steps with step size
$\epsilon_{\mathrm{acc}}/8\simeq1/255$. Each step applies
$\ell_\infty$ projection and image-range clipping. We do not use
random initialization or multiple restarts.

For the association analysis in the main paper, we retain targets
satisfying $s_i(\theta_i^{(u)})<\tau$. For each eligible target with
$i<K$, we measure its maximum future SSCD over the next 30 deletion
stages as
\begin{equation}
m_{i,u}^{(30)}
=
\max_{i<j\leq\min(i+30,K)}
s_i\!\left(\theta_j^{(u)}\right).
\label{eq:app-future-sscd}
\end{equation}
The upper endpoint is truncated at the final checkpoint for targets
near the end of the sequence. A target is labeled as reappearing
when $m_{i,u}^{(30)}\geq\tau$ and as remaining forgotten otherwise.
For the method-level comparison, we average recovery accessibility over
the same $K=300$ deletion targets for SISS and \textsc{LASTING}.

\subsection{Stable Diffusion}
\label{sec:app-sd-protocol}

\paragraph{Data and sequential deletion protocol.}
We implement Stable Diffusion v1.4 using the Hugging Face Diffusers
library. We select $K_{\mathrm{SD}}=30$ deletion targets from the
fully memorized prompt--image pairs identified by Webster and use the
data construction of SISS. Each target contains one reference
memorized image, its original prompt, a modified prompt, and 128
images generated from the modified prompt.

Following SISS, we use target-specific KMeans classifiers to
partition the generated images into memorized and non-memorized
subsets. The memorized subset serves as the stage-specific forget
set, while the non-memorized subset serves as the retain set. Across
the 30 target partitions, this yields 3,840 generated images in total.

We order the 30 targets lexicographically and apply the deletion
requests sequentially. The first stage starts from the pretrained
Stable Diffusion v1.4 pipeline, and the model obtained at stage $i$
initializes stage $i+1$. Each stage uses only the forget and retain
sets of its current target, while the updated model parameters are
carried forward. All methods use the same target order, target
assets, initial model, and checkpoint-transfer rule. We report the
prefix checkpoints at $K\in\{10,20,30\}$.

\paragraph{Memorization and CLIP-IQA evaluation protocol.}
We evaluate memorization separately for the original and modified
prompts and measure generation quality over their combined outputs.
The protocol applies the main-paper persistence metrics to
target--prompt cases using a target-specific KMeans classifier.

\emph{Target-specific memorization classifier.}
We use the precomputed KMeans classifier stored for each target in
the SISS data construction. All evaluation images are generated at
$512\times512$ resolution and passed to the classifier without
resizing. Let $h_i(x)=1$ indicate that the classifier for target $i$
predicts image $x$ as memorized, while $h_i(x)=0$ denotes a
non-memorized image. We keep each classifier fixed across both prompt
types, all evaluation checkpoints, and all compared methods.

\emph{Prompt-level forgetting.}
Immediately after the deletion stage, we evaluate the current target
using its original and modified prompts. At
$K\in\{10,20,30\}$, we reevaluate every target processed through
stage $K$. For target $i$, prompt type
$p\in\mathcal{P}=\{\mathrm{original},\mathrm{modified}\}$, and
checkpoint $\theta_j$, we generate images $x_{i,p,r}^{(j)}$ using
fixed seeds $r\in\{0,\ldots,15\}$. All methods use 50 denoising
steps and a guidance scale of $7.5$. We define the target--prompt
success indicator as
\begin{equation}
z_{i,p}^{(j)}
=
\mathbf{1}\!\left[
\sum_{r=0}^{15} h_i\!\left(x_{i,p,r}^{(j)}\right)=0
\right].
\label{eq:app-sd-prompt-success}
\end{equation}
A target--prompt case is successfully forgotten only when all 16
generations are classified as non-memorized. We apply this condition
independently to the two prompt types, giving two evaluation cases
for each target.

\emph{Target--prompt aggregation.}
At prefix $K$, the $K$ processed targets contribute $2K$
target--prompt cases. We apply the main-paper persistence metrics
to these cases as
\begin{align}
\mathrm{FinalSuccess}(K)
&=
\frac{1}{2K}
\sum_{i=1}^{K}\sum_{p\in\mathcal{P}}z_{i,p}^{(K)},
\label{eq:app-sd-final-success}\\
\mathrm{ReappRate}(K)
&=
\frac{
\sum_{i=1}^{K}\sum_{p\in\mathcal{P}}
\mathbf{1}\!\left[
z_{i,p}^{(i)}=1\land z_{i,p}^{(K)}=0
\right]
}{
\sum_{i=1}^{K}\sum_{p\in\mathcal{P}}z_{i,p}^{(i)}
}.
\label{eq:app-sd-reappearance}
\end{align}
Final Success averages the final forgetting indicators over all
$2K$ cases. The Reappearance Rate denominator contains the cases
successfully forgotten at their immediate checkpoints, and its
numerator counts the same cases that fail the criterion under
checkpoint $\theta_K$.

\emph{CLIP-IQA.}
We measure generation quality using CLIP-IQA. We use the TorchMetrics
\texttt{CLIPImageQualityAssessment} implementation with the
\texttt{clip\_iqa} model, the \texttt{quality} prompt, and data
range $1.0$. The score at each stage averages the 32 images
generated for the current target, comprising 16 images from each
prompt type. Previous-target samples generated at the reported
prefix checkpoints are used only for memorization evaluation and
do not enter the CLIP-IQA average. We apply the same KMeans and
CLIP-IQA evaluation pipeline to all methods, including Prompt-Free.

\paragraph{Compared methods and optimization details.}
We compare \textsc{LASTING} with SISS, EraseDiff, ReTrack, Forward
KL-Constrained Unlearning, and Prompt-Free Instance Unlearning.
Table~\ref{tab:app-sd-optimization} reports the configuration used
for each sequential run.

\begin{table}[ht!]
\caption{Optimization settings for the Stable Diffusion experiments.
Updates denote optimizer updates for each deletion target.}
\label{tab:app-sd-optimization}
\centering
\normalsize
\setlength{\tabcolsep}{2pt}
\renewcommand{\arraystretch}{1.15}
\begin{tabular}{@{}lccccc>{\raggedright\arraybackslash}p{1.65in}@{}}
\toprule
Method
& Optimizer
& \shortstack{Learning\\rate}
& Batch
& Accum.
& \shortstack{Updates\\per target}
& Method-specific setting \\
\midrule
SISS
& AdamW & $1\times10^{-5}$ & 1 & 16 & 35
& $\lambda=0.5$, \texttt{scaling\_norm}$=750$ \\
EraseDiff
& AdamW & $1\times10^{-5}$ & 1 & 16 & 35
& $\eta=1\times10^{-2}$ \\
ReTrack
& AdamW & $1\times10^{-5}$ & 1 & 16 & 30
& $k=10$, $\lambda=0.5$, latent $\ell_2$ distance \\
FU
& Adam & $1\times10^{-5}$ & 1 & 8 & 40
& $\rho=1.25$, dual learning rate $5\times10^{-2}$ \\
Prompt-Free
& AdamW & $1\times10^{-5}$ & 1 & 4 & 60
& $\beta=5\times10^{-5}$, global forget-gradient projection \\
\midrule
\rowcolor{gray!18}
\textsc{LASTING}
& AdamW & $1\times10^{-5}$ & 1 & 16 & 35
& $\epsilon_{\mathrm{nbr}}=0.05$,
  $\eta_{\mathrm{in}}=0.05$, $J=1$ \\
\bottomrule
\end{tabular}
\end{table}

\emph{Common implementation.}
All methods optimize the U-Net while keeping the text encoder and VAE
fixed. We use a constant learning-rate schedule without warmup and
clip the maximum gradient norm at $1.0$. The AdamW runs use
$(\beta_1,\beta_2)=(0.9,0.999)$, weight decay $0.01$, and Adam
epsilon $10^{-8}$. All methods follow the same target order and use
the same stage-specific memorized and non-memorized partitions.

\emph{Baseline optimization.}
EraseDiff matches the retain prediction to diffusion noise and the
forget prediction to uniform random noise. ReTrack constructs
target-specific neighbors from the non-memorized partition using
VAE-latent $\ell_2$ distance and combines its unlearning and retain
losses using the weight reported in
Table~\ref{tab:app-sd-optimization}. Prompt-Free Instance Unlearning
projects the forget gradient when its global inner product with the
retain gradient is negative and then combines it with the retain
gradient.

\emph{FU calibration.}
We calibrate FU on the first deletion target over
$\rho\in\{1.25,1.5,2.0\}$, primal learning rates
$\{5\times10^{-6},1\times10^{-5}\}$, dual learning rates
$\{1\times10^{-2},5\times10^{-2}\}$, and optimizer-update budgets
$\{20,40,60\}$. Eligible candidates achieve a forget-ratio
attainment rate of at least $0.9$ while limiting the relative
retain-loss increase to at most $0.1$. Among these candidates,
we select the configuration with the lowest mean SSCD and use it
for the full sequence. FU applies $\rho$ to the
$\theta_0$-normalized forget-loss ratio. The dual multiplier is
initialized to zero and updated using an EMA coefficient of $0.9$.

\emph{\textsc{LASTING} configuration.}
\textsc{LASTING} retains the SISS outer objective and optimizer.
Its inner search starts from zero perturbation and minimizes the
raw latent-diffusion noise-prediction loss through projected
sign-gradient descent. The inner search and outer deletion update
reuse the same diffusion timestep and Gaussian noise. We detach
the selected latent neighbor before evaluating the SISS deletion
objective.

\emph{Computational environment.}
We run each experiment on a single NVIDIA RTX 5090 GPU with 32 GB
of memory. A complete 30-stage training sequence requires
approximately five to six hours per method.

\section{Additional Qualitative Results}
\subsection{CelebA-HQ}

Figure~\ref{fig:app-celeba-qualitative} presents six additional deletion targets from the CelebA-HQ sequence.
For each target, comparing the immediate and final reconstructions shows how its visual resemblance changes as subsequent deletion requests are processed.
These examples provide target-level context for the aggregate persistence results reported in the main paper.

\newlength{\appqualcell}
\newlength{\appquallabelwidth}
\setlength{\appqualcell}{0.120\textwidth}
\setlength{\appquallabelwidth}{0.090\textwidth}

\newcommand{\appqualimage}[2]{%
  \parbox[c][\appqualcell][c]
    {\dimexpr\appqualcell+1pt\relax}{%
    \centering
    \includegraphics[width=\appqualcell,height=\appqualcell,keepaspectratio]{#1}%
  }%
}

\newcommand{\appqualpair}[4]{%
  \begin{tabular}[c]{@{}c@{}}
    \appqualimage{#1}{#2}\\
    \noalign{\vskip 4pt}
    \appqualimage{#3}{#4}
  \end{tabular}%
}

\newcommand{\appqualrowlabel}[2]{%
  \parbox[c][\appqualcell][c]{\appquallabelwidth}{%
    \centering\scriptsize
    \shortstack{\textbf{#1}\\\textbf{#2}}%
  }%
}

\newcommand{\appqualrowlabels}{%
  \begin{tabular}[c]{@{}c@{}}
    \appqualrowlabel{Immediate}{Model}\\
    \noalign{\vskip 4pt}
    \appqualrowlabel{Final}{Model}
  \end{tabular}%
}

\newcommand{\appqualoriginal}[2]{%
  \parbox[c][\dimexpr2\appqualcell+4pt\relax][c]
    {\dimexpr\appqualcell+1pt\relax}{%
    \centering\appqualimage{#1}{#2}%
  }%
}

\newcommand{\appqualheader}[1]{%
  \parbox[c][2.6em][c]{\appqualcell}{%
    \centering\footnotesize\bfseries #1%
  }%
}

\newcommand{\appqualtarget}[2]{%
  \appqualrowlabels
  &
  \appqualoriginal
    {#1/#2_orig.jpg}
    {#2 Original}
  &
  \appqualpair
    {#1/#2_lasting_imm.png}{#2 Ours Immediate}
    {#1/#2_lasting_final.png}{#2 Ours Final}
  &
  \appqualpair
    {#1/#2_siss_imm.png}{#2 SISS Immediate}
    {#1/#2_siss_final.png}{#2 SISS Final}
  &
  \appqualpair
    {#1/#2_erase_imm.png}{#2 EraseDiff Immediate}
    {#1/#2_erase_final.png}{#2 EraseDiff Final}
  &
  \appqualpair
    {#1/#2_retrack_imm.png}{#2 ReTrack Immediate}
    {#1/#2_retrack_final.png}{#2 ReTrack Final}
  &
  \appqualpair
    {#1/#2_kl_imm.png}{#2 FU Immediate}
    {#1/#2_kl_final.png}{#2 FU Final}
  &
  \appqualpair
    {#1/#2_unpromptable_imm.png}{#2 Prompt-Free Immediate}
    {#1/#2_unpromptable_final.png}{#2 Prompt-Free Final}%
}

\begin{figure}[t]
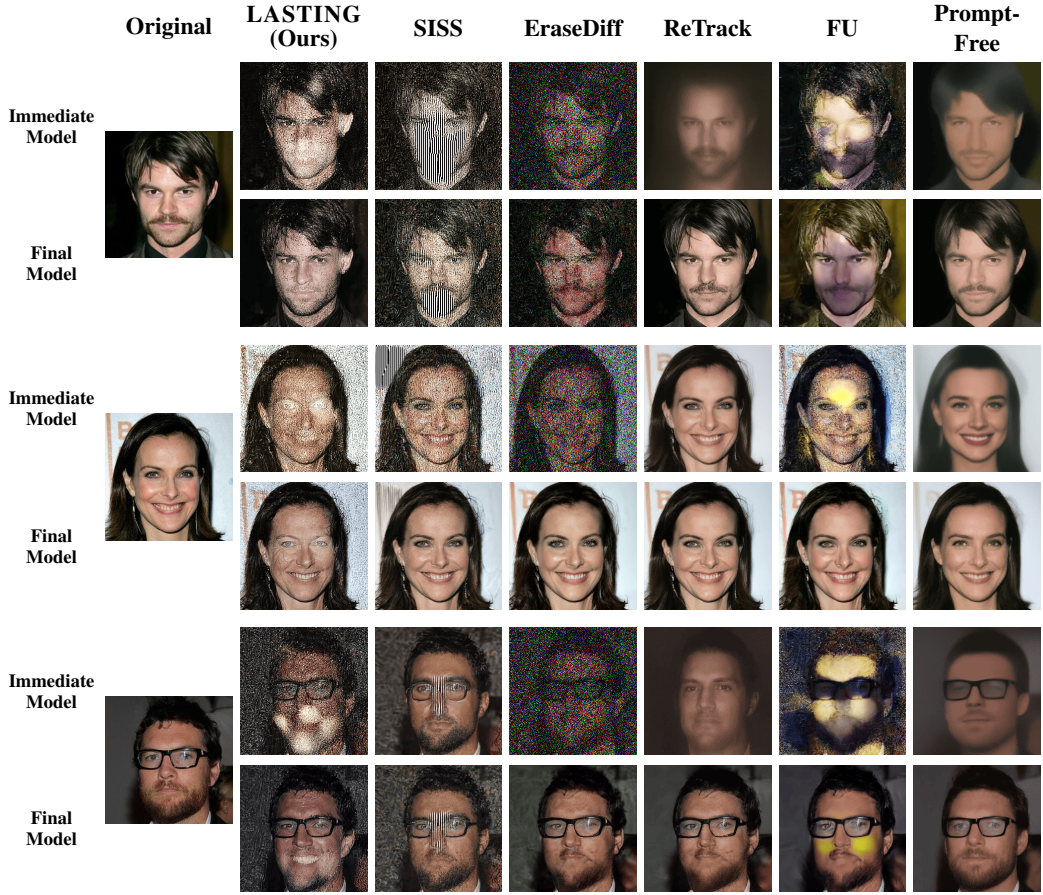

  \centering
  \begin{tabular}{
    @{}c
    @{\hspace{2pt}}c
    @{\hspace{2pt}}c
    @{\hspace{2pt}}c
    @{\hspace{2pt}}c
    @{\hspace{2pt}}c
    @{\hspace{2pt}}c
    @{\hspace{2pt}}c@{}
  }
    &
    \appqualheader{Original}
    &
    \appqualheader{\shortstack{\textsc{LASTING}\\(Ours)}}
    &
    \appqualheader{SISS}
    &
    \appqualheader{EraseDiff}
    &
    \appqualheader{ReTrack}
    &
    \appqualheader{FU}
    &
    \appqualheader{\shortstack{Prompt-\\Free}}
    \\[3pt]

    \appqualtarget{appendix/fig4/10113}{10113}\\
    \noalign{\vskip 7pt}

    \appqualtarget{appendix/fig4/10128}{10128}\\
    \noalign{\vskip 7pt}

    \appqualtarget{appendix/fig4/10148}{10148}
  \end{tabular}

  \caption{Additional qualitative results on CelebA-HQ.
  For each deletion target, the immediate model shows the recovery output
  immediately after that target is unlearned. The final model shows the
  recovery output after all subsequent deletion requests have been processed.}
  \label{fig:app-celeba-qualitative}
\end{figure}

\begin{figure}[t]
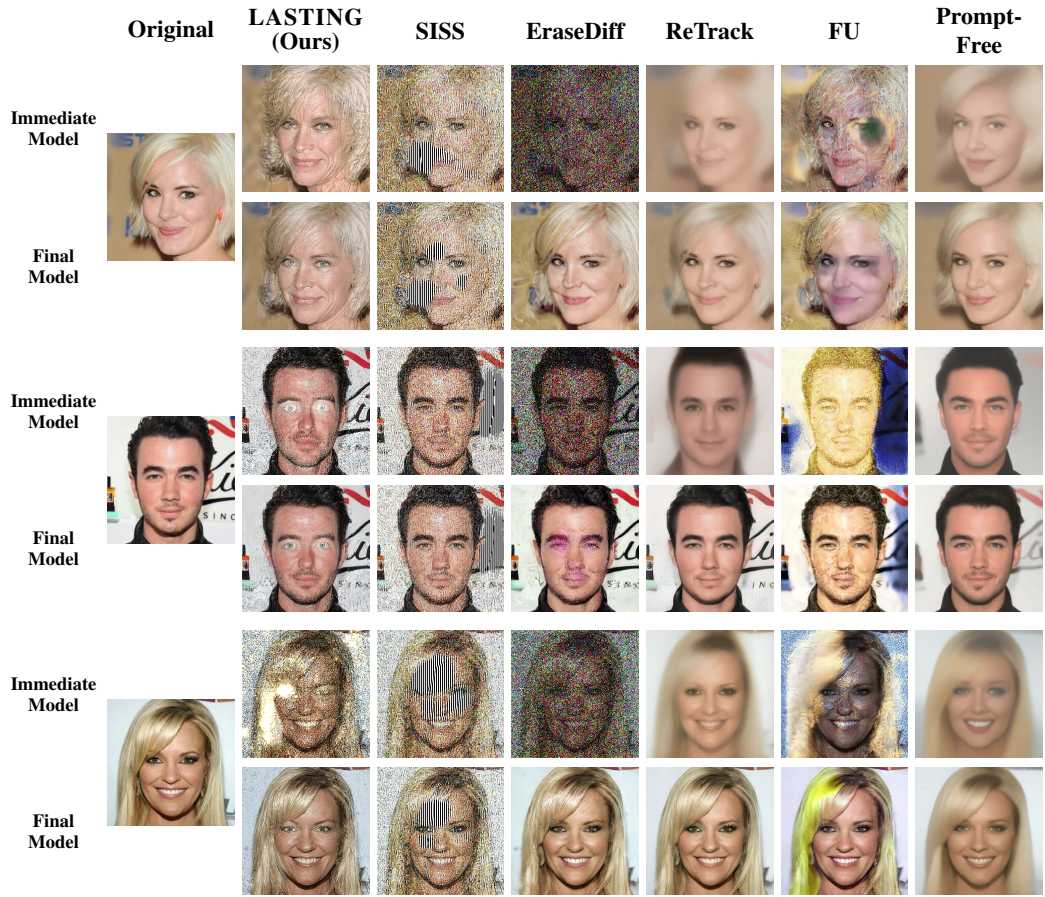

  \ContinuedFloat
  \centering
  \begin{tabular}{
    @{}c
    @{\hspace{2pt}}c
    @{\hspace{2pt}}c
    @{\hspace{2pt}}c
    @{\hspace{2pt}}c
    @{\hspace{2pt}}c
    @{\hspace{2pt}}c
    @{\hspace{2pt}}c@{}
  }
    &
    \appqualheader{Original}
    &
    \appqualheader{\shortstack{\textsc{LASTING}\\(Ours)}}
    &
    \appqualheader{SISS}
    &
    \appqualheader{EraseDiff}
    &
    \appqualheader{ReTrack}
    &
    \appqualheader{FU}
    &
    \appqualheader{\shortstack{Prompt-\\Free}}
    \\[3pt]

    \appqualtarget{appendix/fig4/10162}{10162}\\
    \noalign{\vskip 7pt}

    \appqualtarget{appendix/fig4/10166}{10166}\\
    \noalign{\vskip 7pt}

    \appqualtarget{appendix/fig4/10206}{10206}
  \end{tabular}

  \caption[]{Additional qualitative results on CelebA-HQ (continued).}
\end{figure}

\subsection{Stable Diffusion}

Figure~\ref{fig:app_sd_qualitative} presents four additional memorized prompts and their reference images.
The immediate and final generation composites allow visual inspection of whether reference-specific content remains suppressed after subsequent deletion requests.
Showing four generations per checkpoint also makes variation within each prompt visible.



\newlength{\sdqualcell}
\newlength{\sdquallabelwidth}
\setlength{\sdqualcell}{0.215\textwidth}
\setlength{\sdquallabelwidth}{0.095\textwidth}

\newcommand{\sdqualimage}[1]{%
  \parbox[c][\sdqualcell][c]{\sdqualcell}{%
    \centering
    \includegraphics[
      width=\sdqualcell,
      height=\sdqualcell,
      keepaspectratio
    ]{#1}%
  }%
}

\newcommand{\sdqualpair}[2]{%
  \begin{tabular}[c]{@{}c@{}}
    \sdqualimage{#1}\\[3pt]
    \sdqualimage{#2}
  \end{tabular}%
}

\newcommand{\sdqualrowlabel}[1]{%
  \parbox[c][\sdqualcell][c]{\sdquallabelwidth}{%
    \centering\scriptsize\bfseries #1%
  }%
}

\newcommand{\sdqualrowlabels}{%
  \begin{tabular}[c]{@{}c@{}}
    \sdqualrowlabel{Immediate}\\[3pt]
    \sdqualrowlabel{Final}
  \end{tabular}%
}

\newcommand{\sdqualreference}[1]{%
  \parbox[c][\dimexpr2\sdqualcell+3pt\relax][c]{\sdqualcell}{%
    \centering\sdqualimage{#1}%
  }%
}

\newcommand{\sdqualheader}[1]{%
  \parbox[c][2.6em][c]{\sdqualcell}{%
    \centering\footnotesize\bfseries #1%
  }%
}

\newcommand{\sdqualpromptcontent}[3]{%
  \parbox{\linewidth}{\centering\small\bfseries #3}%
  \par\vspace{5pt}%
  \begin{tabular}{@{}c
    @{\hspace{2pt}}c
    @{\hspace{2pt}}c
    @{\hspace{2pt}}c
    @{\hspace{2pt}}c@{}}
    &
    \sdqualheader{Reference}
    &
    \sdqualheader{\textsc{LASTING}}
    &
    \sdqualheader{SISS}
    &
    \sdqualheader{EraseDiff}
    \\[4pt]
    \sdqualrowlabels
    &
    \sdqualreference{#1/#2_reference.png}
    &
    \sdqualpair
      {#1/#2_lasting_imm.png}
      {#1/#2_lasting_final.png}
    &
    \sdqualpair
      {#1/#2_siss_imm.png}
      {#1/#2_siss_final.png}
    &
    \sdqualpair
      {#1/#2_erasediff_imm.png}
      {#1/#2_erasediff_final.png}
  \end{tabular}%
  \par\vspace{8pt}%
  \begin{tabular}{@{}c
    @{\hspace{2pt}}c
    @{\hspace{2pt}}c
    @{\hspace{2pt}}c
    @{\hspace{2pt}}c@{}}
    &
    \sdqualheader{Reference}
    &
    \sdqualheader{ReTrack}
    &
    \sdqualheader{FU}
    &
    \sdqualheader{Prompt-free}
    \\[4pt]
    \sdqualrowlabels
    &
    \sdqualreference{#1/#2_reference.png}
    &
    \sdqualpair
      {#1/#2_retrack_imm.png}
      {#1/#2_retrack_final.png}
    &
    \sdqualpair
      {#1/#2_fu_imm.png}
      {#1/#2_fu_final.png}
    &
    \sdqualpair
      {#1/#2_promptfree_imm.png}
      {#1/#2_promptfree_final.png}
  \end{tabular}%
}

\begin{figure}[p]
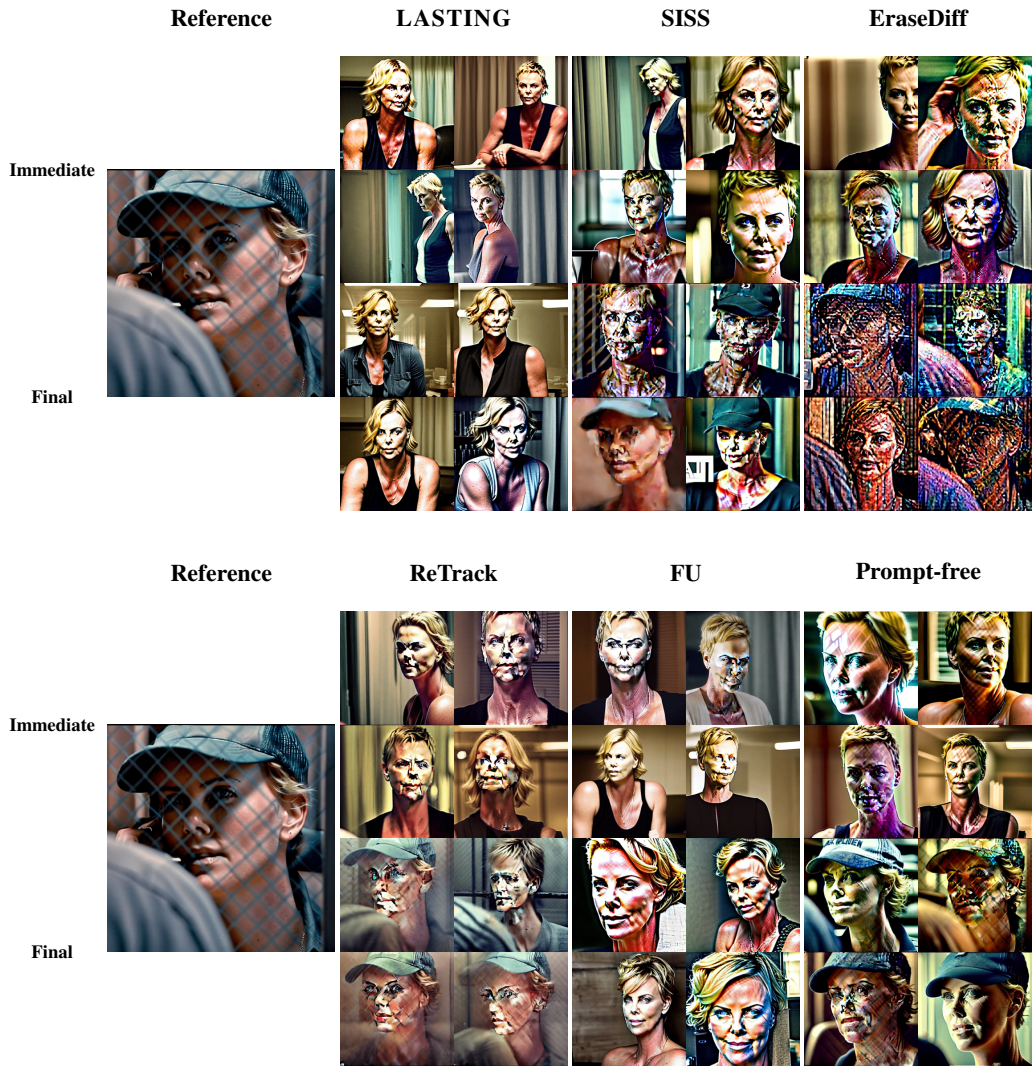

  \centering
  \setlength{\tabcolsep}{0pt}
  \sdqualpromptcontent
    {appendix/fig5/prompt1}
    {prompt1}
    {Original prompt: Video: Charlize Theron in Trailer for
     New Gillian Flynn Adaptation, \emph{Dark Places}}
  \caption{Additional qualitative results on Stable Diffusion.
    Each prompt is divided into two method groups, with the reference
    image repeated in each group. Each method image is a $2\times2$
    composite of four generations. Immediate and final outputs use
    the same prompt and four random seeds across methods.}
  \label{fig:app_sd_qualitative}
\end{figure}

\begin{figure}[p]
  \ContinuedFloat
  \centering
  \setlength{\tabcolsep}{0pt}
  \sdqualpromptcontent
    {appendix/fig5/prompt2}
    {prompt2}
    {Original prompt: Beige on White Watercolor Skull Bedding}
  \caption[]{Additional qualitative results on Stable Diffusion
    (continued).}
\end{figure}

\begin{figure}[p]
  \ContinuedFloat
  \centering
  \setlength{\tabcolsep}{0pt}
  \sdqualpromptcontent
    {appendix/fig5/prompt3}
    {prompt3}
    {Original prompt: Prince Reunites With Warner Brothers,
     Plans New Album}
  \caption[]{Additional qualitative results on Stable Diffusion
    (continued).}
\end{figure}

\begin{figure}[p]
  \ContinuedFloat
  \centering
  \setlength{\tabcolsep}{0pt}
  \sdqualpromptcontent
    {appendix/fig5/prompt4}
    {prompt4}
    {Original prompt: Talks on the Precepts and Buddhist Ethics}
  \caption[]{Additional qualitative results on Stable Diffusion
    (continued).}
\end{figure}

\end{document}

%% file: math_commands.tex
\usepackage{amsmath,amsfonts,bm}

\def\eqref#1{equation~\ref{#1}}

\def\1{\bm{1}}

\DeclareMathAlphabet{\mathsfit}{\encodingdefault}{\sfdefault}{m}{sl}
\SetMathAlphabet{\mathsfit}{bold}{\encodingdefault}{\sfdefault}{bx}{n}

